\documentclass[12pt]{article}
\usepackage{fullpage,graphicx,psfrag,amsmath,amsfonts,verbatim,bm}
\usepackage{amsthm}
\usepackage{amssymb}
\usepackage{algorithm}
\usepackage{algpseudocode}
\usepackage{enumitem}
\usepackage{pgfplots}
\usepackage{linegoal}
\usetikzlibrary{arrows.meta}
\usepackage{amsmath}
\usepackage{kantlipsum}
\usepackage{physics}
\pgfplotsset{compat=1.11}
\allowdisplaybreaks

\newtheorem*{remark}{Remark}

\usepackage[
    backend=biber,
    style=authoryear,
    maxcitenames=1
  ]{biblatex}

\theoremstyle{definition}
\newtheorem{definition}{Definition}

\newtheorem{theorem}{Theorem}
\newtheorem{corollary}{Corollary}
\newtheorem{lemma}{Lemma}
\title{Robust Regression over Averaged Uncertainty}
\author{Dimitris Bertsimas\footnote{Sloan School of Management and Operations Research Center, MIT, Corresponding author, dbertsim@mit.edu} \quad Yu Ma\footnote{Operations Research Center, MIT, midsumer@mit.edu}
}

\begin{document}
\maketitle

\begin{abstract}
\noindent
We propose a new formulation of robust regression by integrating all realizations of the uncertainty set and taking an averaged approach to obtain the optimal solution for the ordinary least squares regression problem. We show that this formulation recovers ridge regression exactly and establishes the missing link between robust optimization and the mean squared error approaches for existing regression problems. We further demonstrate that the condition of this equivalence relies on the geometric properties of the defined uncertainty set. We provide exact, closed-form, in some cases, analytical solutions to the equivalent regularization strength under uncertainty sets induced by $\ell_p$ norm, Schatten $p$-norm, and general polytopes. We then show in synthetic datasets with different levels of uncertainties, a consistent improvement of the averaged formulation over the existing worst-case formulation in out-of-sample performance. In real-world regression problems obtained from UCI datasets, similar improvements are seen in the out-of-sample datasets. 
\end{abstract}


\newpage

\section{Introduction}
Protecting against data uncertainty is at the center of modern machine learning modeling in both the predictive and generative paradigms (\cite{bertsimas2004price, bertsimas2019robust, hastie_09_elements-of.statistical-learning, Hariri2019}). Uncertainties in both the input and outcome data could be attributed to implementation, recording, and manual errors. Examples such as incorrect vital readings during hospital patient stay, as well as manual mistakes on temperature recordings for climate change, are ubiquitous and inherent problems in most real-world applications that can impact the solution quality of the original problem if solved directly. Furthermore, issues such as over-fitting may lead to worse performances in out-of-sample validations if original formulations do not account for uncertainty (\cite{Bhlmann2011, Goodfellow-et-al-2016}). 

\hfill

\noindent
The most prominent approach to address this problem is the use of regularization by incorporating an additional regularizer that either penalizes or encourages certain structures of the solution (\cite{Wang2006, Kratsios2020}). Classical approaches such as lasso and ridge regression have been studied extensively with demonstrated good results in practice. Another approach to account for adversarial noise in the data is by formulating the original least squares problem as a robust optimization problem (\cite{bertsimas2018data, Bertsimas2011, ElGhaoui1997, ro2, ro3, ro4, ro5}). That is, given an uncertainty set that characterizes some belief of the uncertainty in data, we aim to find the optimal solution under the worst-case scenarios. The existing robust optimization formulation offers several advantages. By explicitly defining the adversarial perturbations the model is protecting against, this framework provides additional insights into the behaviors of solutions and beliefs of the original data. It also leads to a more straightforward analysis of the estimators (\cite{ro4}) as well as algorithms for finding the estimators (\cite{BenTal2015}).  

\hfill 

\noindent
There exists a wealth of work that has demonstrated a deeper connection between the robust optimization framework and the regularization approaches, where a main result from (\cite{origina_db}) characterizes the conditions that estbalished the equivalence of robust optimization formulation and lasso. Yet a key observation of this existing approach is that instead of the root-mean-square regression established in these works, in practice, a traditional least squares problem is what is implemented and solved. The least squares formulation offers advantages in computational simplicity since it is closed-form solvable. This curiosity thus begs the natural question of whether there exists a missing link between the traditional robust optimization framework and the current regularization methods. In addition, no computation of exact analytical solutions are available for the regularization strengths even when these least squares cases could be established under other related settings, such as distributionally robust optimization, which could provide insights into the problem settings. 

\hfill

\noindent
In this work, we reformulate the traditional worst-case robust optimization formulation into an averaged approach by accounting for all realizations of the uncertainty set uniformly. By studying the robust linear regression problem under symmetric and non-symmetric uncertainty sets, we provide exact, closed-form, in some cases analytical solutions of the regularization strength. We show that this equivalence relies on geometric properties of the uncertainty set, and demonstrate that when the equivalence holds, these derived solutions achieve better computational performance in both synthetic and real-world data set. 

\subsection{Related Literature}
\subsubsection{Statistical Properties of Ridge Regression}
Ridge regularization has several interpretations that provide insightful statistical properties. One classical interpretation arises from principal components analysis (PCA), where ridge regularization performs shrinkage with more emphasis on the directions corresponding with low variance (\cite{hastie_elements_2009}). This angle implies that ridge regression has the effect of stabilizing solutions by minimizing components with little informational content. Another interpretation is under the setting of a Bayesian framework with normal-normal models. In this context, ridge estimator is shown to be the Bayes estimator when both the prior and the likelihood functions are normal distributions (\cite{10.2307/2987923}). More recent works on high-dimensional statistics also demonstrated ridge's noise protection capacity: we can effectively recover the linear ridge regularization solution if we append a large number of noisy features with zero-mean, unit-variance entries in the original input feature matrix and apply min-norm least squares on this augmented matrix (\cite{10.5555/3455716.3455885}). 

\subsubsection{Equivalence of Robustness and Ridge Regression}
Several recent studies have established the connection between ridge regularization, or even more general regularization techniques, with robustness. Specifically, these works can be classified into three domains of formulation: robust optimization, stochastic optimization and distributionally robust optimization (DRO). Under the lens of robust optimization, where solutions are protected against worst-case scenarios in a deterministic uncertainty set, several works have shown that protection against global noise (or entry-wise) perturbations is equivalent to ridge regression, or more generally, $\ell_p$ norm regression problems. This approach was first established in (\cite{ElGhaoui1997, NIPS2008_24681928}), and then generalized in (\cite{origina_db}) to characterize the exact conditions. In contrast to robust optimization's deterministic nature, stochastic optimization looks for a solution that protects against all realizations of an assumed probability distribution that characterizes the true distribution and thus accounts for distribution information in its formulation. Specifically, previous works have shown that under both additive (\cite{Bishop1995}) and multiplicative (\cite{JMLR:v15:srivastava14a}) stochastic noises, we can recover ridge regularization in neural network settings. Bridging between the two domains and incorporating the advantages of both paradigms, DRO has been proposed as a unifying approach to view the robustification-regularization connection (\cite{Blanchet_Kang_Murthy_2019, ShafieezadehAbadeh2017RegularizationVM, r_paper}). Specifically, DRO identifies solutions that minimize the expected worst-case loss across an ambiguity set, which is formed using empirical distributions and characteristics presumed to represent the true underlying distribution. Several works established the equivalence of lasso linear regression (\cite{JMLR:v19:17-295}), regularized logistic regression for continuous (\cite{ShafieezadehAbadeh2015DistributionallyRL}) and mixed features (\cite{selvi2022wasserstein}). Importantly, these works reveal insightful connections between the size of the Wasserstein balls and the magnitude of regularization strengths.

\subsubsection{Interpretations of Regularization Strength}
Existing works on establishing the equivalence between robustified regression models and regularized regression often arrive at insightful conclusions with respect to the relationship between the defined uncertainty set or ambiguity set and the regularization strength. In (\cite{r_paper}), the regularization for a linear regression case is characterized by the product of the Wasserstein ball's radius, and the Hessian of the loss function (in this case least squares). Similarly in (\cite{ShafieezadehAbadeh2015DistributionallyRL}), the regularization strength for a logistic regression case coincides with the radius of the defined Wasserstein ball.

\hfill 

\noindent
Additional studies on the behaviors of the optimal regularization strength also revealed interesting connections to several factors of the original problem and data setting. (\cite{Dobriban2015HighDimensionalAO}) showed that under appropriate assumptions, the asymptotic optimal regularization strength is a function of both the aspect ratio (ratio of feature size and sample size) and the signal-to-noise (SNR) ratio of the true linear fit. Another interesting angle is provided by the recent observations of the double descent behavior in overparametrized models, predominantly neural networks. Specifically, (\cite{10.5555/3455716.3455885}) showed that the optimal regularization strength can be zero or negative under ill-posed, real-world high-dimensional cases, thus implying that over-parametrization of the model implicitly leads to regularization. These works provide a novel lens into the interpretation and understanding of ridge regression regularization strength.

\subsection{Contributions}
In this paper, we reformulate robust optimization under the worst-case to robust optimization under an averaged uncertainty set, by optimizing the solution over all realizations of the uncertainty set uniformly. We study this formulation for linear regression using both symmetric and non-symmetric uncertainty sets. Our contributions are as follows: 
\begin{itemize}
    \item We provide exact, closed-form, in some cases analytical solutions to the regularization strengths under different conditions of uncertainty sets for linear regression. 
    \item We provide a principled, natural, and theoretical justification for why we should solve the least squares problem under a robust optimization lens in addition to its known computational advantages. 
    \item We demonstrate that the exact equivalence of ridge regression and robust linear regression relies on the geometric properties of the uncertainty set, and show that this equivalence is no longer true under non-symmetric settings. 
    \item We justify the squared formulation as an appropriate model to solve by providing evidence of some of its empirical advantages using both synthetic and real-world datasets.
\end{itemize}

\subsection{Structure of the Paper}
The structure of the paper is as follows: in section 2, we provide an overview of robust optimization and define the uncertainty sets we consider. In section 3, we outline the general characterizations of the robust regression under averaged uncertainty set formulation and its connection to traditional formulations. In section 4, we establish necessary general results over considered uncertainty sets, separated in the symmetric and non-symmetric cases. In section 5, we prove and outline the main theorems demonstrating the new formulation's equivalence with linear ridge regression. In section 6, we demonstrate the experimental results on synthetic and real-world datasets that show the advantage of this formulation over traditional robust optimization. In section 7, we address some concluding remarks.

\section{Brief Overview of Robust Optimization}

\subsection{Norms}
We first introduce the necessary notions of norms: given a vector space $V \subseteq \mathbb{R}^n$, we say that $|| \cdot ||: V \rightarrow \mathbb{R}$ is a \textit{norm} if for all $\mathbf{v}, \mathbf{w} \in V$ and $\alpha \in \mathbb{R}$ we have the following:
\begin{enumerate}
    \item If $|| \mathbf{v}|| = 0$, then $\mathbf{v} = 0$,
    \item $|| \alpha \mathbf{v}|| = |\alpha| ||\mathbf{v}||$ (absolute homogeneity), and
    \item $|| \mathbf{v} + \mathbf{w}|| \leq ||\mathbf{v}|| + ||\mathbf{w}||$ (triangle inequality)
\end{enumerate}

\noindent
Two widely used choices for matrix norms are Frobenius and Schatten norms, which are defined as below.
\begin{enumerate}
    \item The $p$-Frobenius norm, denoted $|| \cdot ||_{F_p}$, is the entrywise $\ell_p$ norm on the entries of $\bm{\Delta} \in \mathbb{R}^{n \times k}$:
    \begin{align*}
    \| \bm{\Delta} \|_{F_p} = \left( \sum_{i=1}^n \sum_{j=1}^k |\bm \Delta_{ij}|^p \right)^{1/p}.
    \end{align*}

    \item The Schatten ($p$-spectal) norm, denoted as $|| \cdot ||_{\mathcal{S}_p}$ is the $\ell_p$ norm on the singular values of the matrix $\bm{\Delta}$:
      \[
\|\bm{\Delta}\|_{\mathcal{S}_p} = \begin{cases} 
\left( \sum_{j=1}^{\min\{n, k\}}\mu_j(\bm{\Delta})^p \right)^{1/p}, &  p < \infty, \\
\max \{\mu_1(\bm{\Delta}), \ldots, \mu_n(\bm{\Delta})\},  &  p = \infty,
\end{cases}
\]
    where $\mu_i(\bm{\Delta})$ denotes the $i$-th entry of the vector containing the singular values of $\bm{\Delta}$.
\end{enumerate}

\subsection{Dual Norms}
The concept of a \emph{dual norm} plays a significant role in the context of robust optimization and is derived from a specific optimization problem that seeks to maximize the linear function \( \mathbf{a}^\top \mathbf{x} \) subject to a norm constraint on \( \mathbf{x} \). Formally, for a given vector \( \mathbf{a} \in \mathbb{R}^n \), the dual norm \( \|\mathbf{a}\|_{q^*} \) is defined as the solution to the following problem:
\[
\max_{\|\mathbf{x}\|_q \leq 1} \mathbf{a}^\top \mathbf{x}.
\]

\noindent
Here, the norm \( \|\mathbf{a}\|_{q^*} \) corresponds to the dual of the \( \ell_q \) norm, where \( q^* \) is the conjugate exponent of \( q \), satisfying \( \frac{1}{q} + \frac{1}{q^*} = 1 \). For instance, when \( q = 2 \), the dual norm is simply the Euclidean norm, while for \( q = 1 \), the dual norm is the \( \ell_\infty \) norm, which represents the maximum absolute value among the components of the vector.

\hfill

\noindent
This duality is further extended to the setting where the vector \( \mathbf{x} \) is scaled by a factor \( \rho \), leading to the modified problem:
\begin{align*}
    \max_{\|\mathbf{x}\|_q \leq \rho} \mathbf{a}^\top \mathbf{x} = \rho \|\mathbf{a}\|_{q^*},
\end{align*}
\noindent
indicating that the solution scales linearly with \( \rho \). The concept of dual norms is also generalizable to matrices, where the dual norm is defined via the trace inner product and is crucial for understanding the behavior of matrix norms in higher dimensions.

\subsection{Robust Optimization}
Robust optimization is a powerful methodology for addressing optimization problems under uncertainty, particularly when the uncertainty is not easily modeled probabilistically. Instead of relying on probability distributions, robust optimization constructs a deterministic \textit{uncertainty set}, denoted by \( \mathcal{U} \), which encapsulates all possible realizations of the uncertain parameters. The goal is to find a solution that remains feasible and optimal across all realizations within \( \mathcal{U} \). Formally, consider an optimization problem where the decision variables \( \mathbf{x} \in \mathcal{X} \) must satisfy a set of constraints defined by a vector-valued function \( \mathbf{g}(\mathbf{u}, \mathbf{x}) \leq \mathbf{0} \) for all \( \mathbf{u} \in \mathcal{U} \). Here, \( \mathcal{X} \subseteq \mathbb{R}^n \) represents the feasible region for \( \mathbf{x} \), and \( \mathbf{u} \in \mathbb{R}^m \) denotes the vector of uncertain parameters. The robust counterpart of the original optimization problem can be formulated as follows:
\[
\max_{\mathbf{x} \in \mathcal{X}} \min_{\mathbf{u} \in \mathcal{U}} c(\mathbf{u}, \mathbf{x}),
\]
\[
\text{subject to } \mathbf{g}(\mathbf{u}, \mathbf{x}) \leq \mathbf{0}, \quad \forall \mathbf{u} \in \mathcal{U},
\]

\noindent
where \( c(\mathbf{u}, \mathbf{x}) \) is the objective function that depends on both the decision variables and the uncertain parameters. The inner minimization problem identifies the worst-case realization of the objective function within the uncertainty set \( \mathcal{U} \), while the outer maximization problem seeks the best possible decision \( \mathbf{x} \) that optimizes the objective under this worst-case scenario.

\hfill

\noindent
Although the robust formulation introduces an infinite number of constraints—corresponding to the infinite possible values of \( \mathbf{u} \) within \( \mathcal{U} \)—it is often possible to reformulate the problem as a finite-dimensional, deterministic optimization problem. This reformulation depends on the specific structure of \( \mathcal{U} \) and the functional forms of \( c(\mathbf{u}, \mathbf{x}) \) and \( \mathbf{g}(\mathbf{u}, \mathbf{x}) \). The resulting deterministic problem, often referred to as the \textit{robust counterpart}, can be solved using conventional optimization techniques. The advantages of robust optimization are well-documented in the literature, particularly in scenarios where small perturbations in the data can lead to significant violations of feasibility or optimality in the nominal solution. By explicitly considering the worst-case scenario, robust solutions provide a higher degree of reliability, thereby ensuring performance that is both stable and resilient to uncertainty.

\subsection{Global-Robustness}
To capture our belief of the structure of the noise we aim to protect against, we construct uncertainty sets that obey certain boundedness conditions. Specifically, in this case, we consider boundedness conditions of the entire noise matrix of the form, or global robustness, where $\rho$ is a parameter controlling the magnitude of the considered perturbations and, hence, the degree to which the features in the training set are able to deviate from their nominal values:
\begin{align*}
    \mathcal{U} = \left\{ \mathbf{\Delta} \in \mathbb{R}^{n \times k} \mid \| \mathbf{\Delta} \| \leq \rho \right\}. 
\end{align*}

\noindent
Some commonly considered global-robustness uncertainty sets are defined as follows using the Frobenius norm:
\begin{itemize}
    \item Ellipsoidal uncertainty set refers to 
    \begin{align}
    \label{ellipsoid-uncertainty}
        \mathcal{U}_1 = \{\bm{\Delta}: \| \bm{\Delta}\|_{F_2} \leq \rho\}
    \end{align}
    \item Box uncertainty set refers to 
    \begin{align}
    \label{box-uncertainty}
    \mathcal{U}_2 = \{\bm{\Delta}: \| \bm{\Delta} \| _{F_{\infty}} \leq \rho\}
    \end{align}
    \item Diamond uncertainty set refers to 
    \begin{align}
    \label{diamond-uncertainty}
    \mathcal{U}_3 = \{\bm{\Delta}: \|\bm{\Delta}\|_{F_1} \leq \rho\}
    \end{align}
    \item Budget uncertainty set refers to 
    \begin{align}
    \label{budget-uncertainty}
    \mathcal{U}_4 = \{\bm{\Delta}: \| \bm{\Delta} \| _{F_1} \leq \Gamma, \| \bm{\Delta}\|_{F_{\infty}} \leq \rho \}
    \end{align}
\end{itemize}

\hfill

\noindent
In addition, we also consider uncertainty sets that are defined by Schatten norm ball and a general polytope, which are defined as follows:
\begin{itemize}
    \item Schatten uncertainty set refers to  
    \begin{align}
    \label{schatten-uncertainty}
    \mathcal{U}_{\mathcal{S}_p} = \{\bm{\Delta}: \| \bm{\Delta} \|_{\mathcal{S}_p} \leq \rho\}
    \end{align}
    \item Polytopal uncertainty set refers to 
    \begin{align}
    \label{polytope-uncertainty}
    \mathcal{U}_{P} = \{\bm{\Delta}: b - A^\top \bm{\Delta} \geq 0\},
    \end{align} where $P$ is a polytope that can be triangularted into $t$ distinct simplices $\bm \Lambda_1, \cdots, \bm \Lambda_t$.
\end{itemize}

\hfill 

\noindent
Lastly, for completeness, we also provide definition of the uncertainty sets that protect against feature and label noise. Given data matrix $\bm{X} = (\bm{x}_1, \cdots, \bm{x}_n) \in \mathbb{R}^{n \times k}$, with the $i$-th data sample and outcmome $\bm{x}_i \in \mathbb{R}^{k} \text{ and } \bm{y}_i$, let $\mathbf{\Delta X} = (\mathbf{\Delta} \bm{x}_1, \mathbf{\Delta} \bm{x}_2, \cdots, \mathbf{\Delta}\bm{x}_n)$. 

\begin{itemize}
    \item The feature-wise uncertainty set is defined as:
    \[
    \mathcal{U}_x = \left\{ \mathbf{\Delta X} \in \mathbb{R}^{n \times k} \mid \| \mathbf{\Delta}\bm{x}_i \| \leq \rho, i = 1, \ldots, n \right\}. 
    \]
    \item The label-wise uncertainty set is defined as:
\begin{itemize}
    \item For binary classification purpose: 
    \begin{align*}
    \mathcal{U}_y = \left\{ \mathbf{\Delta}\bm{y} \in \{-1, 1\}^n \middle| \sum_{i=1}^n \bm{\Delta}\bm{y}_i \leq \rho \right\}. 
\end{align*}
    \item For regression purpose:
    \begin{align*}
    \mathcal{U}_y = \left\{ \mathbf{\Delta}\bm{y} \in \mathbb{R}^{n} \mid \| \mathbf{\Delta}\bm{y}_i \| \leq \rho, i = 1, \ldots, n \right\}. 
\end{align*}
\end{itemize}
\end{itemize}

\section{Robust Optimization under Averaged Uncertainty}

A disadvantage of the existing robust optimization formulation is that the solutions it recovers protect against the worst-case uncertainty of the defined uncertainty set. This approach assumes that the data is under the most severe perturbations and thus arrives at solutions that could be too conservative (\cite{roos2020reducing}). An intuitive remedy is to instead seek a solution that is robust over the averaged realization of uncertainties, thus avoiding over-protecting extreme perturbations.

\subsection{Characterization of Averaged Uncertainty}
We provide the characterization of the new averaged formulation and discuss its connection to stochastic optimization, as well as distributionally robust optimization.

\begin{definition}[RO Average]
Given a data matrix \(\bm{X} \in \mathbb{R}^{n \times k}\), where \(n\) is the number of samples and \(k\) is the number of features and an outcome data vector \(\bm{y} \in \mathbb{R}^n \), the optimal robust optimizer under averaged uncertainty set solution $\bm{\beta}$ is the optimal solution to the following problem. 
\begin{align}
\label{ro-average}
     \min_{\bm{\beta}} \left( \int_{\mathcal{U}} g(\bm{X}, \bm{\Delta}, \bm{y})\, \mathrm{d}\mathcal{U} \right)
\end{align}
\end{definition}

\noindent
Note that this is equivalent to a stochastic optimization problem with uniform distribution over the defined uncertainty set. The computation of the expectation of uniform distribution over a convex polytope has been studied extensively in literature. We choose to adopt this particular robust-optimization-inspired formulation to study the analytical forms of well-known uncertainty sets, and to exploit its deterministic nature leveraging results in numerical analysis. Similarly, in Distributionally Robust Optimization (DRO), we could also consider this formulation as an approximation of an ambiguity set that is as close to the uniform distribution as possible. For example, by considering a sequence of $n$-th order moment constraints that characterize the uniform distribution.

\subsection{Connections to Other Robustness Methods}
We outline some well-known results establishing the equivalence between robustness and regularization, and elaborate on the connection of our formulation with these existing approaches.

\hfill

\noindent It is well-known in the literature the equivalence between general norm-induced robust optimization formulation with $\ell_p$ regression problems of the following. 

\begin{theorem}[\cite{origina_db, Bertsimas2011}]
If $r, q \in [1, \infty]$, and $\mathcal{U}_{(q, r)} = \{\bm{\Delta}: \| \bm{\Delta} \|_{(q, r)} \leq \lambda\}$ with $\| \bm{\Delta} \| _{(q, r)} = \max_{\bm{\beta} \in \mathbb{R}} \frac{\| \bm{\Delta\beta} \|_r}{\| \bm{\beta} \|_q}$ then 
\begin{equation*}
     \min_{\bm{\beta}} \max_{\bm{\Delta} \in \mathcal{U}_{(q, r)}} \| \bm{y} - (\bm{X} + \bm{\Delta}) \bm{\beta} \|_r  = \min_{\bm{\beta}} \| \bm{y} - \bm{X\beta}\|_r + \lambda \| \bm{\beta} \|_q.
\end{equation*}
\end{theorem}

\noindent However, as previously pointed out, one key observation is that this formulation does not solve the true least squares problem, and instead resorts to a root-mean-square problem which is not practically used. We will show in the main result that we bridge this gap by establishing the exact equivalence with the least squares case with RO Average introduced in \ref{ro-average}. 

\hfill

\noindent Another related stream of distributionally robust optimization literature has also established similar equivalence. 

\noindent
Let $\mathbb{S}_{++}^d$ to denote the set of $d$-by-$d$ positive definite matrices, 
$\|X\|_M \triangleq \sqrt{X^\top M X}$ for any $X \in \mathbb{R}^d, M \in \mathbb{S}_{++}^d$, $\delta_X$ denote the Dirac measure at $X$ and let $\widehat{\mathbb{P}} \triangleq \frac{1}{N} \sum_{i=1}^{N} \delta_{X_i}$ be the empirical measure constructed from sample $\{X_1, \dots, X_N\}$. We also define $c(\cdot, \cdot) : \mathbb{R}^d \times \mathbb{R}^d \to [0, \infty)$ as a lower semi-continuous cost function such that $c(X, X) = 0$ for every $X \in \mathbb{R}^d$. We further denote \(\mathcal{P}(\mathcal{X} \times \mathcal{X})\) as the set of joint probability distribution $\pi$ of $(\bar{X}, X)$ supported on $\mathcal{X} \times \mathcal{X}$, while $P_1 \pi$ and $P_2 \pi$ respectively refer to the marginals of $\bar{X}$ and $X$ under the joint distribution $\pi$. Given \(
L_{\beta}(\widehat{\mathbb{P}}, \rho)\) as the worst-case expected loss under all possible distributions around the empirical measure $\widehat{\mathbb{P}}$ at most $\rho$ with respect to the optimal transport distance, and $\rho \geq 0$ as the radius of the uncertainty set centered at \(\widehat{\mathbb{P}} \), then the exact martingale DRO problem is formulated as below,
\begin{equation}
\label{dro-result}
\min_{\beta} L_{\beta}(\widehat{\mathbb{P}}, \rho), \quad
\text{where} \quad
L_{\beta}(\widehat{\mathbb{P}}, \rho) \triangleq 
\left\{
\begin{alignedat}{2}
    &\sup_{\pi} \quad & & \mathbb{E}_{\pi}[\ell(f_{\beta}(\bar{X}))] \\
    &\text{s.t.} & & \pi \in \mathcal{P}(\mathcal{X} \times \mathcal{X}) \\
    & & & \mathbb{E}_{\pi} [c(\bar{X}, X)] \leq \rho, \quad P_2\pi = \widehat{\mathbb{P}} \\
    & & & \mathbb{E}_{\pi}[\bar{X} | X] = X, \quad \widehat{\mathbb{P}}\text{-a.s.}, \
\end{alignedat}
\right. 
\end{equation}
\noindent 
which gives the following equivalence result.

\begin{theorem}
Suppose that (i) the loss function $\ell(\cdot)$ is a convex quadratic function, i.e., $\nabla^2 \ell(\cdot) = \gamma > 0$, and (ii) the feature mapping $f_\beta(\mathbf{X}) = \beta^\top \mathbf{X}$ is linear.  Let $X^\top \triangleq (Y, Z^\top) \in \mathbb{R}^d$ and $\beta^\top \triangleq (1, -b^\top) \in \mathbb{R}^d$, we have $\beta^\top X = Y - b^\top Z$. For any $Q \in \mathbb{S}^{d-1}_{++}$, we take $M = \text{diag}(+\infty, Q)$, then the problem (\ref{dro-result}) with $\gamma = 2$ becomes
\[
\min_{b} \left\{ \mathbb{E}_{\hat{\mathbb{P}}} \left[(Y - b^\top Z)^2 \right] + \rho \|b\|_{Q^{-1}}^2 \right\}.
\]
\end{theorem}

\noindent
This DRO formulation, under certain conditions, yields an exact equivalence with the traditional least squares approach. In comparison, our approach takes an alternative deterministic perspective. Notably, we demonstrate the conditions where this equivalence holds, and show that when the uncertainty set lacks symmetry, the conditions for this equivalence will be violated. Furthermore, we provide closed-form, and in some cases analytical, solutions for the regularization strength term across various conventional uncertainty sets.

\section{General Results over Uncertainty Sets}
In this section, we provide and prove useful conclusions of some of the most commonly used uncertainty sets in robust optimization. We note that these conclusions are special cases of the broader topic of the study of convex bodies. However, leveraging the specific boundedness conditions and unique geometric properties of these chosen uncertainty sets, we are able to derive exact, closed-form, in some cases analytical solutions, that can provide insights into the equivalence between robust optimization and ridge regularization. For all of the following, we consider the setting where $\bm \Delta \in \mathbb{R}^{n \times k}$, and $\mathcal{U} = \left\{ \mathbf{\Delta} \in \mathbb{R}^{n \times k} \mid  \text{some boundedness condition} \right\}$. 



\hfill

\noindent
The rest of the discussion is separated into two distinct classes of convex bodies based on their geometric structure: symmetric or non-symmetric, which we will later see drive some key observations in our formulation.

\subsection{Symmetric Uncertainty Sets}
We outline below results on the zeroth-, first-, and second-order functions under these symmetric settings, respectively corresponding to the volume, specialized odd function, and quadratic functions. 

\subsubsection{Zeroth-Order Functions: Volume}
The zeroth-order function, primarily concerned with the volume of symmetric uncertainty sets, serves as a basic measure of their \textit{size} or \textit{capacity}. Understanding volume is crucial because it directly affects the feasibility region of optimization problems—larger volumes imply greater uncertainty but also potentially higher robustness against data variability. Results concerning volumes of common symmetric sets, such as cubes and hyperspheres, establish metrics that can guide the selection and application of these sets in practical scenarios.
 
\begin{lemma}
\noindent
For the most commonly used $\ell_p$-norm based uncertainty sets, their volumes in high dimensions ($\mathbb{R}^n$ below) are as follows:
\begin{enumerate}
    \item Hypercube (or box uncertainty set): a hypercube with side length $a$ has volume $V = a^n$,
    \item Hypersphere (or spherical uncertainty set): a hypersphere with radius $a$ has volume $V = \frac{\pi^{n/2}}{\Gamma(n/2 + 1)}a^n$,
    \item Simplex: a simplex with vertices at the origin and unit vectors along the axes has volume $V = \frac{1}{n!}$,
    \item Ellipsoid (or ellipsoidal uncertainty set): an ellipsoid defined by $\frac{x_1^2}{a_1^2} + \frac{x_2^2}{a_2^2} + \cdots + \frac{x_n^2}{a_n^2} \leq 1$ has volume $V = \frac{\pi^{n/2}}{\Gamma(n/2 + 1)}(a_1 a_2 \cdots a_n)$.
\end{enumerate}
\end{lemma}

\begin{lemma}[\cite{coxeter}]
\label{coxeter}
    Let $V_{\mathcal{U}_3}$ denote the hypervolume of the diamond uncertainty set defined in (\ref{diamond-uncertainty}), or sometimes also referred to as the hyper cross-polytope, then \[V_{\mathcal{U}_3} = \frac{(2\rho)^{nk}}{(nk)!}.\]
\end{lemma}

\begin{lemma}
    Let $V_{\mathcal{U}_4}$ denote the hypervolume of the budget uncertainty set defined by (\ref{budget-uncertainty}), then, \[V_{\mathcal{U}_4} = \frac{(2\rho)^{nk} - nk(2(\rho - \Gamma))^{nk}}{(nk)!}.\] 
    \begin{proof}
\noindent
 $V_{\mathcal{U}_4}$ is the volume of a polytope defined by the region $A = \{\bm{\Delta} \in \mathbb{R}^{n \times k}: \|\bm{\Delta}\|_{F_1} \leq \rho\}$ truncated out by $2nk$ corners of the regions  $\{\bm{\Delta} \in \mathbb{R}^{n \times k}: \|\bm{\Delta}\|_{F_1} \leq \rho, \|\bm{\Delta}\|_{F_{\infty}} > \Gamma \}$. From Lemma \ref{coxeter}, the volume of $A$ is $\frac{(2\rho)^{nk}}{(nk)!}$ and two of these $2nk$ corners from opposing sides can be combined into a polytope of volume $\frac{2(\rho - \Gamma)^{nk}}{(nk)!}$, and thus we have
$V_{\mathcal{U}_4} = \frac{(2\rho)^{nk}}{(nk)!} -\frac{nk(2(\rho - \Gamma))^{nk}}{(nk)!}$. \qedhere 
\end{proof}
\end{lemma}

\noindent
Another class of uncertainty sets that should be considered is those defined by the Schatten norm, which is one of the most important classes of unitary operators and has a long sequence of literature investigating its behavior using asymptotic geometric analysis. 

\begin{theorem}[\cite{KABLUCHKO2020105457}]
We provide below the asymptotic volume of Schatten norm ball. Given $A$ as a $n \times n$ matrix with entries from $\mathbb{R}$, $\mathcal{S}_p$ denoting the Schatten $p$-norm. If we denote by \( 
B_p^n(\mathbb{R}) = \{ A: \|A\|_{\mathcal{S}_p} \leq 1\}
\) the corresponding Schatten unit ball, and Vol$_N$ the Lebesgue measure of dimension \(N \in \mathbb{N}\), we have that as as \( n \rightarrow \infty \), 

\[
\left(\text{Vol}_{n^2} B_p^n(\mathbb{R})\right)^{1/n^2} \sim n^{-\frac{1}{2} - \frac{1}{p}} \sqrt{2\pi e^{3/2}\sigma(p/2)}, 
\]
where, 
\[ 
\sigma(p) = \frac{1}{4} \left( \frac{2\sqrt{\pi}\Gamma(p+1)}{\sqrt{e}\Gamma(p+\frac{1}{2})} \right)^{1/p}. 
    \]
\end{theorem}

\begin{remark}
We note that the result on the Schatten norm differs from previous $\ell_p$-norm balls since only asymptotic results can be established. In addition, we note that existing results can only be applied to square matrices instead of a more general $n \times k$ matrix.
\end{remark}

\subsubsection{First Order Functions}
Due to the symmetric nature of the norm-induced uncertainty sets we consider, we introduce some useful general results first on odd functions. 
\begin{definition}
    We define the set $\mathcal{U}  \subset \mathbb{R}^n$ as a symmetric set about the origin if for every matrix $\bm{\Delta} \in \mathcal{U}$, the matrix $-\bm{\Delta}$ also belongs to $\mathcal{U}$. In other words, $\bm{\Delta} \in \mathcal{U}$ implies $-\bm{\Delta} \in \mathcal{U}$. 
\end{definition}

\begin{lemma}[Symmetry of Norm-Based Uncertainty Sets]
It is immediately obvious that the uncertainty sets defined by global robustness using both $\ell_p$-norm and Schatten $p$-norm are symmetric sets. 
\end{lemma}

\begin{lemma}[Univariate Symmetry]
If $\mathcal{U} \subset \mathbb{R}$ is a symmetric interval around the origin and $f(x)$ is an odd function, then \[\int_{\mathcal{U}} f(x) \, \dd x = 0.\]
\begin{proof}
    Since $\mathcal{U}$ is symmetric, for every $x \in \mathcal{U}, -x \in \mathcal{U}$ as well. Since $f$ is odd, $f(-x) = -f(x)$. By changing variables in the integral, we have:
    $\int_{\mathcal{U}} f(x) \, \dd x = \int_{\mathcal{U}} f(-y) \, \dd y = -\int_{\mathcal{U}} f(y) \, \dd y$. We thus conclude that $\int_{\mathcal{U}} f(x) \, \dd x = 0$. 
\end{proof}
\end{lemma}

\begin{corollary}[Multivariate Symmetry]
\label{multivariate-symmetry}
Let $\mathcal{U} \subset \mathbb{R}^n$ be a symmetric set around the origin and $f(\bm x): \mathbb{R}^n \rightarrow \mathbb{R}$ be an odd function with respect to each of its variables, then \[\int_{\mathcal{U}} f(\bm x) \, \dd x = 0. \]
\begin{proof}
    Since $\mathcal{U}$ is symmetric about the origin, for every $\bm x \in \mathcal{U}, -\bm x \in \mathcal{U}$ as well. Since $f$ is odd, \(f(x_1, x_2, \cdots, -x_i, \cdots, x_n) = -f(x_1, x_2, \cdots, x_i, \cdots, x_n)\) for all $i = 1, \cdots, n$. By applying univariate symmetry with respect to each dimension of $\bm{x}$, we arrive at the conclusion. 
\end{proof}
\end{corollary}


\begin{corollary}[Matrix Symmetry]
\label{matrix-symmetry}
    If $\mathcal{U} \subset \mathbb{R}^{n \times k}$ is symmetric around the origin and $f(v)$ is a function independent of $\mathbf{\Delta}$, and $g(\mathbf{\Delta}): \mathbb{R}^{n \times k} \rightarrow \mathbb{R}^{n \times k}$ is an odd function, then \[\int_{\mathcal{U}} f(v) g(\mathbf{\Delta}) \, \dd \mathbf{\Delta} =\bm{0}, \] 
    where $\mathbf{0}$ is a matrix of the same dimension of $\mathbf{\Delta}$ with all entries of 0. 
    \begin{proof}
    Applying Corollary \ref{multivariate-symmetry} to each entry of the matrix integral yields the result. 
    \end{proof}
\end{corollary}

\begin{corollary}
    It is immediately obvious that when $\mathcal{U}$ is the global-robustness uncertainty sets previously defined using $\ell_p$ norm and Schatten $p$-norm and an odd function $g(\mathbf{\Delta})$ we have: \[\int_{\mathcal{U}} g(\mathbf{\Delta}) \, \dd \mathbf{\Delta} = \mathbf{0}.\]
\end{corollary}

\subsubsection{Quadratic Functions}
Ridge regression is defined as a quadratic function, and we establish some related results for general symmetric sets. 
\begin{lemma} 
\label{u1-lemma1}
If $\bm{\Delta} \in \mathbb{R}^{n \times k}$ and $V(nk, \rho)$ is the volume of $\mathcal{U}_1$ defined in (\ref{ellipsoid-uncertainty}), then:
\begin{align*}
\int_{\mathcal{U}_1} \bm{\Delta}^\top \bm{\Delta} \, \dd \mathbf{\Delta} = 
\begin{bmatrix}
\frac{V(nk, \rho)}{k} & 0 & \cdots & 0 \\
0 & \frac{V(nk, \rho)}{k} & \cdots &  0 \\
\cdots & \cdots & \cdots & \cdots \\
0 & 0 & \cdots & \frac{V(nk, \rho)}{k}\\
\end{bmatrix}.
\end{align*}
\begin{proof}
    Please see Appendix section \ref{u1-proof}. 
\end{proof}
\end{lemma}

\begin{lemma} 
\label{u2-lemma1}
If $\bm{\Delta} \in \mathbb{R}^{n \times k}$, and $\mathcal{U}_2$ is defined in (\ref{box-uncertainty}), then:
\begin{align*}
\int_{\mathcal{U}_2} \bm{\Delta}^\top \bm{\Delta} \, \dd \mathbf{\Delta} = 
\begin{bmatrix}
\frac{(2\rho)^{nk}\rho^2 n}{3} & 0 & \cdots & 0 \\
0 & \frac{(2\rho)^{nk}\rho^2 n}{3} & \cdots &  0 \\
\cdots & \cdots & \cdots & \cdots \\
0 & 0 & \cdots & \frac{(2\rho)^{nk}\rho^2 n}{3}\\
\end{bmatrix}. 
\end{align*}
\begin{proof}
    Please see Appendix section \ref{u2-proof}. 
\end{proof}
\end{lemma}

\begin{lemma}
\label{u3-lemma1}
If $\bm{\Delta} \in \mathbb{R}^{n \times k}$, and $\mathcal{U}_3$ is defined in (\ref{diamond-uncertainty}), then 
\begin{align*}
\int_{\mathcal{U}_3} \bm{\Delta}^\top \bm{\Delta} \, \dd \mathbf{\Delta} = 
\begin{bmatrix}
\frac{(2\rho)^{nk+1}\rho n}{(nk+2)!} & 0 & \cdots & 0 \\
0 & \frac{(2\rho)^{nk+1}\rho n}{(nk+2)!}  & \cdots &  0 \\
\cdots & \cdots & \cdots & \cdots \\
0 & 0 & \cdots & \frac{(2\rho)^{nk+1}\rho n}{(nk+2)!} \\
\end{bmatrix}.
\end{align*}
\begin{proof}
    Please see Appendix section \ref{u3-proof}. 
\end{proof}
\end{lemma}

\begin{lemma}
\label{u4-lemma1}
If $\bm{\Delta} \in \mathbb{R}^{n \times k}$, and $\mathcal{U}_4$ is defined in (\ref{budget-uncertainty}), then 
\begin{align*}
\int_{\mathcal{U}_4} \bm{\Delta}^\top \bm{\Delta} \, \dd \mathbf{\Delta} = 
\begin{bmatrix}
f(n, k, \rho, \Gamma) & 0 & \cdots & 0 \\
0 & f(n, k, \rho, \Gamma) & \cdots &  0 \\
\cdots & \cdots & \cdots & \cdots \\
0 & 0 & \cdots & f(n, k, \rho, \Gamma) \\
\end{bmatrix}, 
\end{align*}
where $f(n, k, \rho, \Gamma) = \frac{2\rho^2}{(n+1)(n+2)}\frac{(2\rho)^n - n(2(\rho - \Gamma))^n}{n!} - \frac{(2(\rho - \Gamma))^n}{n!}\frac{(n^2+3n-2)\Gamma^2 + (4-2n)\rho \Gamma}{(n+1)(n+2)}$. 
\begin{proof}
    Please see Appendix section \ref{u4-proof}. 
\end{proof}
\end{lemma}

\hfill 

\noindent
Similarly to the study of volume, we resort to the literature on asymptotic geometric analysis for the Schatten norm ball. 
\begin{definition}
    A compact, convex subset \(K\) with a non-empty interior is called a convex body. Furthermore, it is called \textit{isotropic} if (i) its Lebesgue volume \(\text{vol}(K) = 1\), (ii) it is \textit{centered}, that is, has a barycentre at the origin, and (iii) its covariance matrix is a multiple of the identity, namely
\[
\int_K x_i x_j \, \dd x = L_K^2 \mathbf{1}_{ij} \quad \text{for all } 1 \leq i, j \leq m,
\]
\(L_K\) here is called the \textit{isotropic constant} of \(K\), \(\mathbf{1}_{ij}\) is the indicator function indicating 1 if $i = j$, and 0 otherwise. 
\end{definition}

\begin{definition}
    We define $\mathcal{N}$ as a \textit{unitarily invariant} norm on the space \(\mathcal{M}_n(\mathbb{R})\) with $n \times n$ matrices with real entries if it satisfies \(\mathcal{N}(USV) = \mathcal{N}(S)\) for any \(S \in \mathcal{M}_n(\mathbb{R})\)  and any real isometries \(U, V\) on \(\mathbb{R}^n\) with the Euclidean norm. 
\end{definition}

\begin{lemma}[\cite{Knig1998}]
    The Schatten $p$-norm is a unitarily invariant norm. 
\end{lemma}

\begin{lemma}[\cite{Knig1998}]
     The unit balls $B_{\mathbb{R}}(\mathcal{N})$ of a unitarily invariant normed space of matrices (with norm $\mathcal{N}$) are isotropic.
\end{lemma}

\noindent
The above two conclusions lead immediately to the following.
\begin{corollary}
    The Schatten unit balls $B_{\mathbb{R}}(\mathcal{S}_p^n) = \{A \in \mathcal{M}_n(\mathbb{R}); \bm s_p(A) \leq 1\}$ is isotropic.
\end{corollary}

\begin{theorem}[\cite{Knig1998}] The isotropic constant of Schatten class is bounded. Let us denote by $L_{\mathbb{R}}(n,p) = L^2_{B_{\mathbb{R}}(\mathcal{S}_p^n)}$ where $L_{B_{\mathbb{R}}(\mathcal{S}_p^n)}$ is the isotropic constant of the Schatten $p$-norm ball. We have \[\quad L_{\mathbb{R}}(n,p) \simeq n^{-\frac{2}{p}} \frac{M_p(x_1^2)}{M_p(1)}, \] where $M_p$ is the measure with density \[f_{n,p}(x_1, \dots, x_n) = \mathbf{1}_{\{x_1 \geq 0, \dots, x_n \geq 0\}} f_n(x) e^{-\sum_{i=1}^n x_i^p}, \] with respect to the Lebesgue measure on $\mathbb{R}^n$.
\end{theorem}

\subsection{Non-Symmetric Uncertainty Sets}
We consider the case of an $n$-dimensional polytope $P$ defined by a union of simplices $\bm \Lambda_i$, and observe that there are exact, closed-formed solutions for expressing integrals of arbitrary polynomials under certain conditions. 




\begin{theorem}[\cite{baldoni2011integrate}]
\label{baldoni}
    Let \( \bm \Lambda \) be the simplex that is the convex hull of \( \mathbf{s}_0, \mathbf{s}_2, \dots, \mathbf{s}_{d} \) in \( \mathbb{R}^n \), and let \( \ell \) be an arbitrary linear form on \( \mathbb{R}^n \). Then
\[
\int_{\bm \Lambda} \ell^M \, \dd m = d! \, \text{vol}(\bm \Lambda) \, \frac{M!}{(M+d)!} \sum_{\mathbf{k} \in \mathbb{N}^{d+1}, |\mathbf{k}| = M} \langle \ell, \mathbf{s}_0 \rangle^{k_1} \dots \langle \ell, \mathbf{s}_{d} \rangle^{k_{d+1}}, 
\]
where \( |\mathbf{k}| = \sum_{j=1}^{d+1} k_j \).
\end{theorem}

\begin{corollary}
\label{first-order-polytope}
        Setting $M$ = 1, we immediately have the following result. Let \( \bm \Lambda \) be the simplex that is the convex hull of \( \mathbf{s}_0, \mathbf{s}_2, \dots, \mathbf{s}_{d} \) in \( \mathbb{R}^n \), $\bm{x} \in \mathbb{R}^n$ a point in the simplex, and $\bm{x}_i$ the $i$-th index of the point. We denote $\{\mathbf{s}_{j}\}_i$ as the value of the $i$-th entry of the vector $\mathbf{s}_{j}$, and $\dd m$ is the integral Lebesgue measure, then:
    \[
    \int_{\bm \Lambda} \bm{x}_i \, \dd m = \frac{\text{vol}(\bm \Lambda)}{d + 1} \sum_{j = 0}^d \{\mathbf{s}_{j}\}_i.
    \]
\end{corollary}

\begin{corollary}
\label{cor-polytope-first}
    Let $P$ be a polytope that can be triangulated into $t$ simplices $\bm \Lambda_1, \bm \Lambda_2, \cdots, \bm \Lambda_t$, where each simplex $\bm \Lambda_{\kappa}$ is defined by its vertices \( \mathbf{s}_{0 \kappa}, \mathbf{s}_{2 \kappa}, \dots, \mathbf{s}_{d \kappa} \), then:
    \[
    \int_P \bm{x}_i \, \dd m = \sum_{\kappa = 1}^t \frac{\text{vol}\bm \Lambda_{\kappa}}{d+1} \sum_{j = 0}^d \{\mathbf{s}_{j\kappa} \}_i.
    \]
    \begin{proof}
        This immediately follows that \(P = \bigcup_{\kappa = 1}^t \bm \Lambda_{\kappa} \), which leads to \( \int_P \bm{x}_i \, \dd m = \sum_{\kappa = 1}^t \left( \int_{\bm \Lambda_{\kappa}} \bm{x}_i \, \dd m \right)\), and the conclusion follows from Lemma \ref{first-order-polytope}. 
    \end{proof}
\end{corollary}

\begin{corollary}
    Let \(\bm \Lambda \) be the simplex that is the convex hull of \( \mathbf{s}_0, \mathbf{s}_2, \dots, \mathbf{s}_{d} \) in \( \mathbb{R}^n \), $\bm{x} \in \mathbb{R}^n$ a point in the simplex, and $\bm{x}_i$ the $i$-th index of the point, then:
\[
\int_{\bm \Lambda} \bm{x}_i^2 \, \dd m = \frac{2 \times  \text{vol}(\bm \Lambda)}{(d+2)(d+1)} \left( \sum_{j = 0}^{d} \{\mathbf{s}_j\}_i^2 + \sum_{j \neq r} 2 \{\mathbf{s}_j\}_i \{\mathbf{s}_r\}_i \right).\]
\begin{proof}
     Let \( \ell \) be an arbitrary linear form on \( \mathbb{R}^n \). After setting $M = 2$ in Theorem \ref{baldoni}, cancelling the factorials, we observe that to satisfy \( |\mathbf{k}| = \sum_{j=0}^{d} k_j = 2\text{ for }\mathbf{k} \in \mathbb{N}^{d+1}\), entries of $k_j$ can only be either 1 or 2, then:
    \[
\int_{\bm \Lambda} \ell^2 \, \dd m = \frac{2 \times  \text{vol}(\bm \Lambda)}{(d+2)(d+1)} \left( \sum_{j = 0}^{d} \langle \ell, \mathbf{s}_j \rangle^2 + \sum_{r \neq j} 2 \langle \ell, \mathbf{s}_i \rangle \langle \ell, \mathbf{s}_j \rangle \right).\] The result follows once we apply the appropriate linear form $\ell_i$ for each $i$ (1 for the $i$-th entry and 0 otherwise). 
\end{proof}
\end{corollary}

\begin{corollary}
\label{cor-polytope-second}
It is immediately obvious that given $P$ be a polytope that can be triangulated into $t$ simplices $\bm \Lambda_1, \bm \Lambda_2, \cdots, \bm \Lambda_t$, where each simplex $\bm \Lambda_{\kappa}$ is defined by its vertices $\mathbf{s}_{0\kappa}, \mathbf{s}_{1\kappa}, \cdots, \mathbf{s}_{d\kappa}$. Then we have that:
    \[
    \int_P \bm{x}_i^2 \, \dd m = \sum_{\kappa = 1}^t \frac{2 \times  \text{vol}(\bm \Lambda_{\kappa})}{(d+2)(d+1)} \left( \sum_{j = 0}^{d} \{\mathbf{s}_j\}_i^2 + \sum_{r \neq j} 2 \{\mathbf{s}_j\}_i \{\mathbf{s}_r\}_i \right).\]    
\end{corollary}

\section{Linear Regression}
\subsection{Problem Setting}
Linear regression models the relationship between multiple continuous independent variables and a continuous dependent variable. Given $n$ as the number of samples, $k$ as the number of features, we define the input data $\bm{X} \in \mathbb{R}^{n \times k}$, outcome data $\bm{y} \in \mathbb{R}^n$, and $\bm \beta \in \mathbb{R}^k$ as the desired solution. The linear regression assumes the relationship \( \bm y = \bm X^\top \bm \beta \). The \textit{Ordinary Least Squares (OLS)} minimize the sum of squared residuals \[\min_{\boldsymbol{\beta}} \|\mathbf{y} - \mathbf{X} \boldsymbol{\beta}\|_2^2.\]

\noindent
Incorporating regularization gives the following formulation, where $\lambda$ is the regularization strength usually found through cross-validation.
\begin{equation*}
    \min_{\boldsymbol{\beta}} \|\mathbf{y} - \mathbf{X} \boldsymbol{\beta} \|_2^2 + \lambda g(\bm{\beta}).
\end{equation*}
When $g(\bm{a}) = \|\bm{a}\|_2^2$ and $h(\bm{a}) = \|\bm{a}\|_2^2$, we recover regularized least squares (RLS), or ridge regression (\cite{Hoerl1}). Ridge regression is particularly useful to mitigate the problem of multicollinearity, or highly correlated independent variables, in problems with a large number of parameters. It has also been shown that ridge regression provides a smaller variance and mean square estimator (\cite{RePEc:mtp:titles:026261183x}). Another frequently used regularization approach is $g(\bm{a}) = \|\bm{a}\|_2^2$ and $h(\bm{a}) = \|\bm{a}\|_1$, where we recover the least absolute shrinkage and selection operator, or lasso (\cite{tibshirani96regression}). It is widely believed that the use of lasso can encourage sparsity in the coefficients, (i.e., only a small subset of features coefficients are nonzero) (\cite{Natarajan1995, Tibshirani2005}). Lasso is also computationally efficient since there exist many efficient algorithms to solve it (\cite{Bento2018}). 

\subsection{Taylor Expansion Representation}
We use $n$-th order Taylor expansion of the loss function as a generalized method for evaluating the integration over different uncertainty sets. Specifically, we give the following characterizations of the mean squared loss for linear regression. 

\begin{lemma}
    Let \( f(\bm \Delta) = \|\mathbf{y} - (\mathbf{X} + \bm \Delta) \bm \beta\|_2^2 \), where \( \mathbf{X} \in \mathbb{R}^{n \times k} \) the data matrix, \( \mathbf{y} \in \mathbb{R}^n \) the response vector, \( \bm \beta \in \mathbb{R}^k \)the coefficients, and \( \bm{\Delta} \in \mathbb{R}^{n \times k} \) a perturbation of the data matrix. The function \( f(\bm \Delta) \) can be expressed \textit{exactly} by its second-order Taylor expansion around the zero matrices \(\bm{\Delta} = \bm{0} \) as follows:
    \[f(\bm{\Delta}) = \|\mathbf{y} - \mathbf{X}\bm \beta\|_2^2 - 2(\mathbf{y} - \mathbf{X}\bm \beta)^\top \bm \beta^\top \bm{\Delta} + \bm \beta^\top \bm{\Delta}^\top \bm{\Delta} \bm \beta.\]
    \begin{proof}
        This conclusion follows naturally from Taylor expansion with respective first and second derivatives. We also note that starting from the third derivative, the derivative terms equal to 0 and vanish. However, Taylor expansion does not apply to general functions of matrices trivially, so we derive below the exact formulation. To approximate a general function $F$ to the first order around some matrix $\bm{\Delta}^0$, Taylor's formula gives:
        \[
        f(\bm{\Delta}) = f(\bm{\Delta}^0) + \dd f(\bm{\Delta}^0)(\bm{\Delta} - \bm{\Delta}^0) + \frac{1}{2} \dd^2 f(\bm{\Delta}^0) (\bm{\Delta} - \bm{\Delta}^0, \bm{\Delta} - \bm{\Delta}^0)
        \]

        \noindent
        We have the first term of: 
        \(
        f(\bm{\Delta}^0) = \| \bm{y} - \bm{X\beta} \|^2_2.\\
        \)
        
        \noindent
        To compute the second term \(\dd f(\bm{\Delta^0})\), where \(f(\bm{\Delta}) = \langle \bm{y} - \bm{X\beta} - \bm{\Delta\beta}, \bm{y} - \bm{X\beta} - \bm{\Delta\beta} \rangle \) we use the generalized Leibniz rule and obtain: 
        \[
        \dd f(\bm{\Delta}^0) = \langle \dd(\bm{y} - \bm{X\beta} - \bm{\Delta\beta})(\bm{\Delta}^0) , \bm{y} - \bm{X\beta} - \bm{\Delta}^0 \bm{\beta} \rangle + \langle \bm{y} - \bm{X\beta} - \bm{\Delta}^0 \bm{\beta}, \dd(\bm{y} - \bm{X\beta} - \bm{\Delta\beta})(\bm{\Delta}^0) \rangle.
        \]

        \noindent
        Since the differential of a linear map is the linear map itself, 
        \[
        \dd(\bm{y} - \bm{X\beta} - \bm{\Delta}\bm{\beta})(\bm{\Delta}^0) = -\bm{\beta},
        \]

        \noindent
        and putting together,
        \[
        \dd f(\bm{\Delta}^0) = \langle -\bm{\beta} , \bm{y} - \bm{X\beta} - \bm{\Delta}^0 \bm{\beta} \rangle + \langle \bm{y} - \bm{X\beta} - \bm{\Delta}^0 \bm{\beta}, -\bm{\beta} \rangle \\
        = -2(\bm{y} - \bm{X\beta} - \bm{\Delta}^0 \bm{\beta})^\top \bm{\beta}^\top.
        \]

        \noindent
        For the third term, similarly apply the Leibniz rule again, we have that
        \[
        \dd^2 f(\bm{\Delta}^0) = -2 \langle \bm{\beta}, -\bm{\beta} \rangle.
        \]

        \noindent
        With everything together and setting $\bm{\Delta}^0 = \bm{0}$, we have the conclusion. 
    \end{proof}
\end{lemma}

\begin{remark}
    Note that the conclusion could be easily obtained from standard linear algebra expansions, we adopt the Taylor form to ensure its generalizability for other losses that do not have inherent similarly convenient properties.  
\end{remark}

\subsection{Equivalence with Linear Ridge Regression}
Below we outline the main results that established the equivalence between robust linear regression under averaged uncertainty sets with ridge regression under different uncertainty set settings. We note that the different geometric structures of these uncertainty sets eventually correspond to different strengths and structures of ridge regularization.

\begin{theorem}[$\ell_p$-norm induced ridge regression]
\label{lp-linear-regression-main}
Given a data matrix \(\bm{X} \in \mathbb{R}^{n \times k}\), where \(n\) is the number of samples and \(k\) is the number of features and an outcome data vector \(\bm{y} \in \mathbb{R}^n\), data perturbation matrix $\bm{\Delta} \in \mathbb{R}^{n \times k}$ and $\bm{\beta} \in \mathbb{R}^k$, robust regression under averaged uncertainty is equivalent with ridge regression, 
\begin{center}
    \(\min_{\bm{\beta}} \left(\int_{\mathcal{U}} \| \bm{y} – (\bm{X} + \bm{\Delta}) \bm{\beta} \| _2^2 \, \mathrm{d}\mathcal{U} \right) = \min_{\bm{\beta}} \| \bm{y} – \bm{X\beta} \|^2_2 + \lambda \| \bm{\beta} \|^2_2\).
\end{center}
\begin{enumerate}[label=\alph*)]
    \item For $\mathcal{U} = \mathcal{U}_1$,  the ellipsoidal uncertainty set defined in (\ref{ellipsoid-uncertainty}), $  \lambda = \frac{1}{k}$,
    \item For $\mathcal{U} = \mathcal{U}_2$, the box uncertainty set defined in (\ref{box-uncertainty}), $\lambda = \frac{n\rho^2}{3}$,
    \item For $\mathcal{U} = \mathcal{U}_3$,  the diamond uncertainty set defined in (\ref{diamond-uncertainty}), $\lambda = \frac{2n\rho^2}{(nk+2)(nk+1)}$,
    \item For $\mathcal{U} = \mathcal{U}_4$, the budget uncertainty set defined in (\ref{budget-uncertainty}), $\lambda = \frac{2n\rho^2}{(n+1)(n+2)} - \frac{n(\rho - \Gamma)^n ((n^2 + 3n - 2)\Gamma^2 + (4-2n)\rho\Gamma)}{(n+1)(n+2)((\rho^n - (\rho - \Gamma)^n)}$.   
\end{enumerate}

\begin{proof} We first note that the following general setup follow for all norm-induced global robustness uncertainty sets.     
\begin{align*}
    & \min_{\bm{\beta}} \int_{\mathcal{U}} \| \bm{y} - (\bm{X} + \bm{\Delta}) \bm{\beta} \|_2^2 \, \mathrm{d}\mathbf{\Delta} \\
    &= \min_{\bm{\beta}} \left(\int_{\mathcal{U}} \| \bm{y} - \bm{X\beta} \|_2^2 \, \mathrm{d}\mathbf{\Delta}- 2 (\bm{y} - \bm{X\beta})^\top \bm{\beta} \left(\int_{\mathcal{U}} \bm{\Delta} \, \mathrm{d}\mathbf{\Delta}\right) + \bm{\beta}^\top \left(\int_{\mathcal{U}} \bm{\Delta}^\top \bm{\Delta} \, \mathrm{d}\mathbf{\Delta} \right) \bm{\beta}
    \right) \\
    &= \min_{\bm{\beta}} \left( \text{vol}(\mathcal{U}) \| \bm{y} - \bm{X\beta} \|_2^2 + \bm{\beta}^\top \left(\int_{\mathcal{U}} \bm{\Delta}^\top \bm{\Delta} \, \mathrm{d}\mathbf{\Delta} \right) \bm{\beta} \right) \\ 
     &= \min_{\bm{\beta}} \left( \| \bm{y} - \bm{X\beta} \|_2^2 + 
     \frac{\left(\int_{\mathcal{U}} \bm{\Delta}^\top \bm{\Delta} \, \mathrm{d}\mathbf{\Delta} \right)}{\text{vol}(\mathcal{U})} \bm{\beta}^\top  \bm{\beta} \right) \\
     &= \min_{\bm{\beta}} \left( \| \bm{y} - \bm{X\beta} \|_2^2 + 
    \lambda \| \bm{\beta} \|^2_2 \right).
\end{align*}
where the second step follows Corollary \ref{matrix-symmetry} since the $\ell_p$ norm-induced global robustness uncertainty sets are symmetric around the origin. 

\hfill

\noindent
For ellipsoidal uncertainty sets $\mathcal{U}_1$, given the volume as $V(nk, \rho)$ and $\lambda = \frac{V(nk, \rho)}{k}$. 

\hfill

\noindent
For box uncertainty sets $\mathcal{U}_2$, we have the volume of $(2\rho)^{nk}$ since the volume of a hypercube with a dimension of $nk$ and a side length of $2\rho$ is $(2\rho)^{nk}$. It follows that $\lambda =\frac{(2\rho)^nk \rho^2 n}{3}{k}$.

\hfill

\noindent
For diamond uncertainty set $\mathcal{U}_3$, given the volume $\text{vol}(\mathcal{U}_3) = \frac{(2\rho)^nk }{nk!}$ and $\lambda =\frac{(2\rho)^{nk+1} \rho n}{(nk+2)!}$.

\hfill

\noindent
For the budget uncertainty set, 
\[\lambda = 
f(n, k, \Gamma, \rho) = \frac{2n\rho^2}{(n+1)(n+2)} - \frac{n(\rho - \Gamma)^n ((n^2 + 3n - 2)\Gamma^2 + (4-2n)\rho\Gamma)}{(n+1)(n+2)(\rho^n - (\rho - \Gamma)^n)}.
\]

\hfill 

\noindent
Note that when constructing the budget uncertainty set, our calculation would only be meaningful if $ \sqrt{2}/2 \rho \leq \Gamma \leq \rho$, since if $\Gamma < \sqrt{2}/2 \rho$, we reduce this to the $\| \bm{x} \|_1 \leq \rho$ case, and if $\Gamma > \rho$, we reduce to the $\| \bm{x} \|_{\infty} \leq \Gamma$ case. Thus, given $\Gamma = k\rho$ where $k$ is the scaling factor, the second term above can be simplified to the following:
\begin{align*}
    \frac{n\rho^2(1-k)^n ((n^2 + 3n - 2)k^2 + (4-2n)k)}{(n+2)(n+1)(1-(1-k)^n)}.
\end{align*}
Observe that this term is dominated by the term of $\frac{(1-k)^n}{1-(1-k)^n}$, given reasonable values of $k$ and sufficient sample size (larger than 100), this term will converge to 0 and not dominate the overall constant.
\end{proof}
\end{theorem}

\hfill

\noindent
We further extend this result to Schatten-norm-induced uncertainty sets and establish similar equivalence. 

\begin{theorem}[Schatten-norm induced ridge regression]
    Given a data matrix \(\bm{X} \in \mathbb{R}^{n \times n}\), where \(n\) is the number of samples and \(k\) the number of features and an outcome data vector \(\bm{y} \in \mathbb{R}^n\), data perturbation matrix $\bm{\Delta} \in \mathbb{R}^{n \times n}$, $\bm{\beta} \in \mathbb{R}^{n}$, and $\mathcal{U}_{\mathcal{S}_p}$ as the uncertainty set defined by the Schatten $p$-norm ball in (\ref{schatten-uncertainty}). As the dimension of norm ball $n \rightarrow \infty$,  
\[\min_{\bm{\beta}} \left(\int_{\mathcal{U}_{\mathcal{S}_p}} \| \bm{y} – (\bm{X} + \bm{\Delta}) \bm{\beta} \| _2^2 \, \mathrm{d}\mathbf{\Delta} \right) = \min_{\bm{\beta}} \| \bm{y} – \bm{X\beta} \|^2_2 + \frac{\sqrt{2\pi e^{3/2} \sigma(p/2)} M_p(x_1^2)}{n^{\frac{1}{2} + \frac{2}{p}} M_p(1)} \| \bm{\beta} \|^2_2, \]

\noindent
where, 
\[ 
\sigma(p) = \frac{1}{4} \left( \frac{2\sqrt{\pi}\Gamma(p+1)}{\sqrt{e}\Gamma(p+\frac{1}{2})} \right)^{1/p}, 
    \] and 
$M_p$ is the measure with density \[f_{n,p}(x_1, \dots, x_n) = \mathbf{1}_{\{x_1 \geq 0, \dots, x_n \geq 0\}} f_n(x) e^{-\sum_{i=1}^n x_i^p}, \] with respect to the Lebesgue measure on $\mathbb{R}^n$.
\begin{proof}
    This is a direct application of the volume of the respectively defined Schatten norm ball as well as the result on isotropic constant.  
\end{proof}
\end{theorem}

\hfill

\noindent
Lastly, we show that the equivalence between ridge regression and robust optimization under averaged uncertainty no longer holds under non-symmetric, general polytopal uncertainty sets.

\begin{theorem}[Non-symmetric polytopal protection]
Given a data matrix \(\bm{X} \in \mathbb{R}^{n \times n}\), where \(n\) is the number of samples and $k$ the number of features and an outcome data vector \(\bm{y} \in \mathbb{R}^n\), data perturbation matrix $\bm{\Delta} \in \mathbb{R}^{n \times k}$, $\bm{\beta} \in \mathbb{R}^k$ and $\mathcal{U}_{P}$ as the uncertainty set defined by the polytope $P$ which can be triangulated into $\kappa$ simplices $\mathbf{\Lambda}_1, \cdots, \mathbf{\Lambda}_{\kappa}$ defined in (\ref{polytope-uncertainty}).

\begin{align*}
& \min_{\bm{\beta}} \int_{\mathcal{U}_P }\| \bm{y} – (\bm{X} + \bm{\Delta}) \bm{\beta} \| _2^2 \, \mathrm{d}\mathbf{\Delta} \\
= & \min_{\bm{\beta}} \text{vol}(\mathcal{U}_P) \| \bm{y} - \bm{X\beta} \|_2^2 + 2 (\bm{y} - \bm{X\beta})^\top \bm{\beta} \left( \sum_{\kappa = 1}^m \frac{\text{vol}\bm \Lambda_{\kappa}}{d+1} \sum_{j = 0}^d \{\bm{s}_j \}_i \right)\\
    & + \left( \sum_{\kappa = 1}^t \frac{2 \times  \text{vol}\bm \Lambda_{\kappa}}{(d+2)(d+1)} \left( \sum_{j = 1}^{d} \{\mathbf{s}_j\}_i^2 + \sum_{j \neq r} 2 \{\mathbf{s}_j\}_i \{\mathbf{s}_r\}_i \right)
    \right) \| \bm{\beta} \|^2_2 . 
\end{align*}

\begin{proof}
    \begin{align*}
    & \min_{\bm{\beta}} \int_{\mathcal{U}_P} \| \bm{y} - (\bm{X} + \bm{\Delta}) \bm{\beta} \|_2^2 \, \mathrm{d}\mathbf{\Delta}\\
    &= \min_{\bm{\beta}} \left(\int_{\mathcal{U}_P} \| \bm{y} - \bm{X\beta} \|_2^2 \, \mathrm{d}\mathbf{\Delta} - 2 (\bm{y} - \bm{X\beta})^\top \bm{\beta} \left(\int_{\mathcal{U}_P} \bm{\Delta} \, \mathrm{d}\mathbf{\Delta} \right) + \bm{\beta}^\top \left(\int_{\mathcal{U}_P} \bm{\Delta}^\top \bm{\Delta} \, \mathrm{d}\mathbf{\Delta} \right) \bm{\beta}
    \right). 
\end{align*}
We then apply Corollary \ref{cor-polytope-first} and \ref{cor-polytope-second} on the last equation and arrive at the conclusion. 
\end{proof}
\end{theorem}

\hfill

\noindent
These results establish the important connection between existing robust optimization and least squares ridge regression, where it is the first attempt to bridge a theoretical justification from this perspective for ridge regression. Note that across all symmetric uncertainty sets we consider, the final characterizations all arrive at ridge regression, but with different leading regularization strengths. This implies that ridge regression is a  general regularization method that protects against global perturbations of different noise structures defined under symmetric settings. In addition, we also note that in the more general, non-symmetric, polytopal uncertainty set setting, we no longer recover ridge regression, but with close approximation that accounts for the additional perturbations along the feature-wise axes.

\section{Computational Results}
In this section, we study the performance of averaged uncertainty robust regression (AUR) against worst-case uncertainty robust regression (WUR) using both synthetic and real-world data and found that AUR outperforms WUR across all datasets. All experiments are run using Gurobi 0.11.5, Julia 1.9.3, and Python 3.10.6 using a Mac Intel i7 core. The Homogenous Barrier algorithm was used for the optimization formulation to avoid numerical issues. Our codebase is publicly available for those interested in reproducing results presented in this paper (\cite{Github}). 

\subsection{Computational Experiment Set-up}
The main goal of the experiment is to compare AUR against WUR, defined as follows:
\begin{itemize}
    \item Worst-case uncertainty robust regression (WUR): $\min\limits_{\bm{\beta}} \|\bm{y} - \bm{X\beta}\|_2 + \lambda \|\bm{\beta}\|_2$,  
    \item Averaged uncertainty robust regression (AUR): $\min\limits_{\bm{\beta}} \|\bm{y} - \bm{X\beta}\|_2^2 + \lambda \|\bm{\beta}\|_2^2$. 
\end{itemize}

\noindent
An important remark lies in the two approaches' different objective function formulations. This could cause inconsistencies if we abide by their original forms during the selection of regularization strength $\lambda$. To avoid this issue, we instead apply mean squared error (MSE) for both formulations. To select the regularization strength, we adopt analytical formulas previously derived, or the optimal value selected by cross-validation (CV), which retrieves the best performance on the validation loss. The CV grids are defined by choice of $\lambda$ that ranges from 0 to 1 with 0.05 increments for all experiments to ensure a fine-grain grid for comparison. 

\subsection{Real-World Data}
We selected ten publicly available UCI regression datasets (\cite{Dua:2019}) to analyze the performance of AUR. When missing data is present in the original data, we drop the entire sample. If a feature contains more than 20\% missing values, we drop this feature. We also pre-process the datasets by removing features that do not contain useful information. The final dataset is then standardized using min-max scaling. The information on the datasets is summarized in Table \ref{table1}. To simulate different real-world noises, we added perturbations generated using the hit-and-run (\cite{hit_run}) method from the ellipsoidal, box, diamond, and budget uncertainty sets with values of $\rho \in [0.001, 0.01, 0.05, 0.1, 0.2, 0.3]$. For each dataset and each perturbation strength $\rho$, 10 perturbations are generated using different random seeds to ensure our results account for a diverse perturbation of noise under the same condition. We then split each dataset into 80/20 training and testing sets and applied AUR and WUR respectively to study their out-of-sample MSE performances. Overall, we conducted 2400 experiments that vary across 4 of the $\ell_p$ norm induced uncertainty sets, 10 datasets, 6 perturbation strengths, and 10 perturbation randomness.

\begin{table}[ht!]
\centering
\begin{tabular}{ |c||c|c|}
 \hline
 Dataset Name & Number of Samples & Number of Features\\
 \hline
 Abalone & 4177 & 9\\
 Auto-MPG & 398 & 8 \\
Automobile & 193 & 25 \\
Breast Cancer Wisconsin & 194 & 34 \\
Computer Hardware & 209 & 9 \\
Concrete & 1030 & 9\\
Wine Quality (red) & 1599 & 12 \\
Wine Quality (white) & 4898 & 12 \\
Energy Efficiency & 768 & 9 \\
Synchronous Machine & 557 & 5 \\
 \hline
\end{tabular}
\caption{UCI datasets used in real-world experiments, where sample sizes and feature sizes range different scales.}
\label{table1}
\end{table}

\subsection{Performance on Real World Data}
\begin{figure}[hbt]
\centering
\includegraphics[width=.8\textwidth]{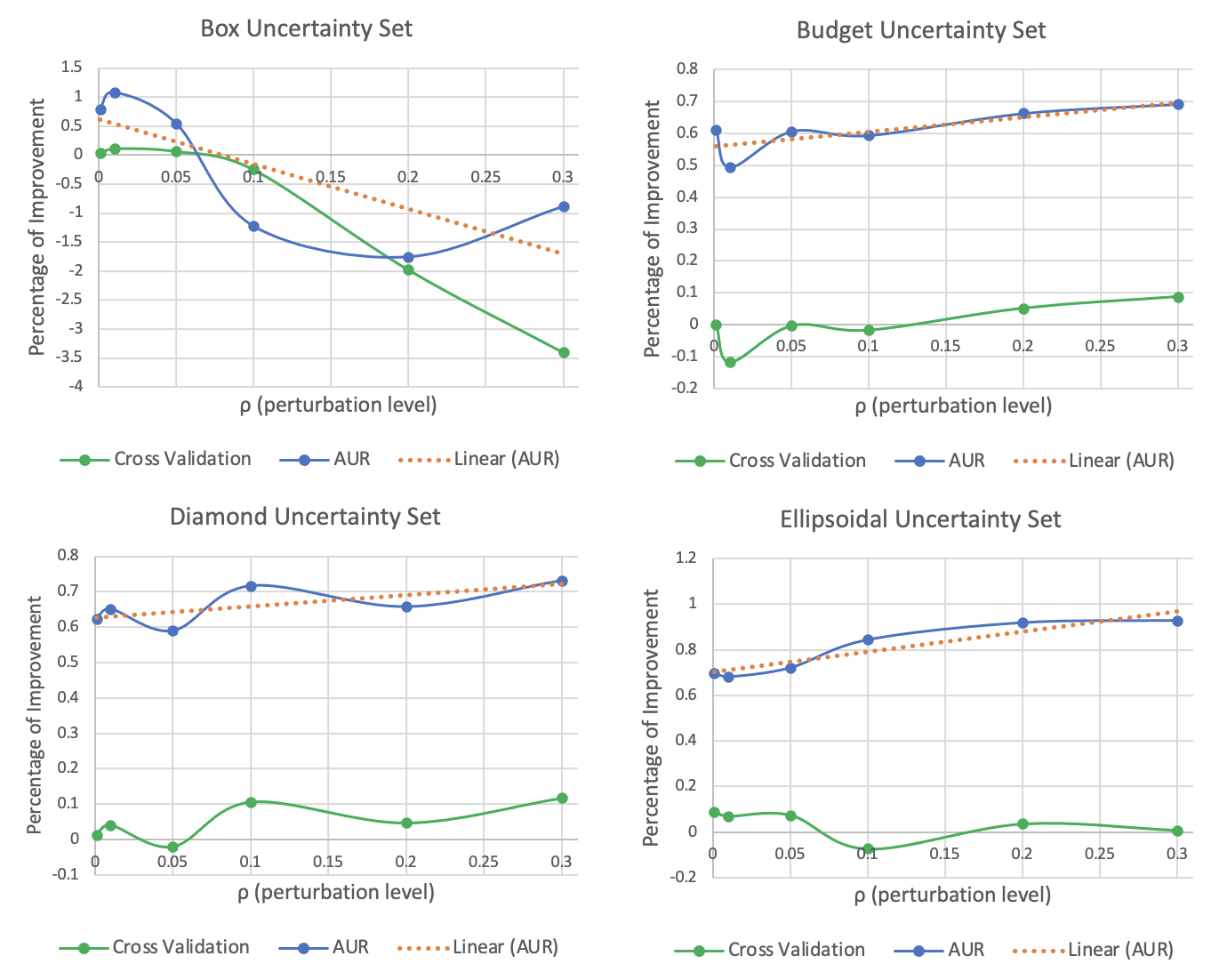}
\caption{Percentage of AUR over WUR across 10 UCI datasets, where the orange line is the trend line for AUC improvements from
 Theorem \ref{lp-linear-regression-main} computed regularization strength. }
\label{Fig3}
\end{figure}

Below we report the MSE of AUR over WUR on the real-world datasets over different uncertainty sets in Figure \ref{Fig3}. We observe that across all different perturbation levels, AUR outperforms WUR by 0.4\% - 0.9\% on average, with improvements increasing as perturbation increases. This result confirms our belief that AUR is able to protect against noise more holistically than traditional WUR, and can be especially useful for real-world datasets when there is strong noise perturbation. We note an exception of the box uncertainty set, which decreases as perturbation increases. We argue that this is because box uncertainty set by nature protects against worst-case global perturbation of every data entry, and is inherently an over-protection. We obtain high $\lambda$ values as the size of the sample size grows, which over-regularizes the training and attributes to this behavior.  

\hfill 

\noindent
Another important observation is the advantage of using regularization strengths obtained by Theorem  \ref{lp-linear-regression-main} in comparison to those obtained by CV. As seen in Figure \ref{Fig3}, we achieve a 0.6-0.8\% MSE improvement. Their improvements are relatively equivalent across different perturbation levels across budget, diamond, and ellipsoidal uncertainty sets, confirming a consistent advantage. 

\begin{table}[ht!]
\centering
\begin{tabular}{ |p{3cm}||p{2cm}|p{2cm}|p{2cm}|p{2cm}|}
 \hline
 $\#$ of Different $\lambda$& Box & Budget & Diamond & Ellipsoidal \\
 \hline
 1 & 23.3\% & 100\% & 100\% & 80.0\%\\
 2 & 23.3\% & 0\% & 0\% & 15.0\%\\
 3 & 23.3\% & 0\% & 0\% & 3.33\%\\
 4 & 15.0\% & 0\% & 0\% & 1.67\%\\
 5 & 10.0\% & 0\% & 0\% & 0\%\\
 6 & 5.0\% & 0\% & 0\% & 0\%\\
 \hline
\end{tabular}
\caption{Frequency of experiments that have different regularization strengths ($\lambda$) obtained with 10 different perturbation noise of the same UCI dataset with the same perturbation strength. It demonstrates the instability of CV regularization strength selection}
\label{table2}
\end{table}

\hfill 

\noindent
Besides the performance improvement offered by using the regularization strengths computed according to Theorem \ref{lp-linear-regression-main}, we also observe that CV is susceptible to the randomness of the training procedure when choosing the optimal regularization strength. Given the same UCI dataset as well as the same uncertainty set, we expect to see the same regularization strengths selected as they protect against the same set of noises. However, we show in Table \ref{table2} that the number of different regularization strengths selected for the same dataset can be as large as 6 using CV, whereas we only need to consider one regularization strength using Theorem \ref{lp-linear-regression-main}. This implies that CV is not the most reliable methodology for computing regularization strengths, as it provides unstable selections as we are exposed to randomness. We note that budget and diamond uncertainty sets give consistent CV-selected regularization strengths, and this is because in practice, when the dimension of the problem becomes large, in order for the perturbation to be contained within the diamond and budget uncertainty sets, the scale of noise becomes smaller than those contained in ellipsoidal or box uncertainty sets, and randomness has a diminished effect on the regularization strength selection.

\subsection{Synthetic Data}
We study more closely the behavior of AUR in comparison to WUR as the number of informative features and the number of samples vary using synthetic datasets. We generated synthetic regression datasets where the regression target is a random linear combination of random features that are well-defined, centered, and unbiased. We vary the number of data samples (300, 400, 500, 600, 700, 900), and a number of informative features (3, 4, 5, 6, 7, 8, 10) to study the effects of these factors on the performance. Additive perturbations are then generated using the hit-and-run method to simulate noise from the ellipsoidal, box, diamond, and budget uncertainty sets. 

\hfill 

\noindent
Specifically, we test the noise level of the uncertainty set of $\rho \in [0.001, 0.01, 0.05, 0.1, 0.2, 0.3]$, where for the budget uncertainty set, we choose $\Gamma = 0.8\rho$. For our samples to truly reflect the monotonically increasing perturbation level, we enforce that samples generated from a higher perturbation level must not reside in the space from the previously smaller perturbation level (i.e., generated perturbation matrix from $\rho = 0.3$ cannot reside in the uncertainty set defined by $\rho = 0.2$). 
To achieve stability of our results,  we repeated each experiment 20 times with a different random seed. 

\subsection{Performance on Synthetic Data}
We observe that AUR improves over WUR across all uncertainty sets, sample sizes, as well as number of informative features. This is in accordance with what we have seen in the real-world datasets. Importantly, the improvements across all uncertainty sets decrease as the number of samples increases, and as the number of informative features increases as shown in Figure \ref{Fig4}. This observation implies that AUR's advantage diminishes as the scale and complexity of the regression problem increases. 

\begin{figure}[hbt]
    \centering
    \includegraphics[width=.7\textwidth]{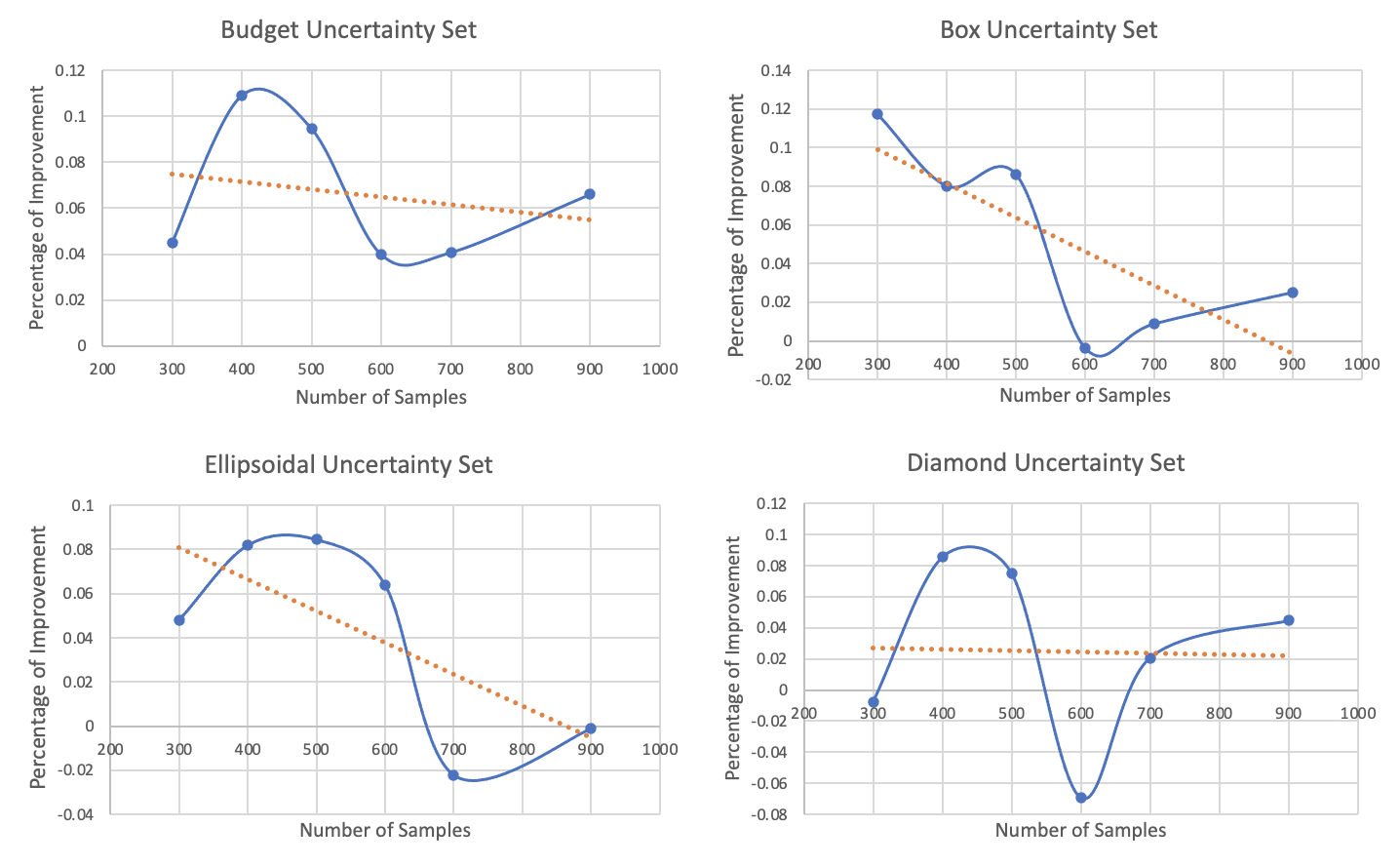}
\end{figure}
    
\begin{figure}
    \centering
    \includegraphics[width=.7\textwidth]{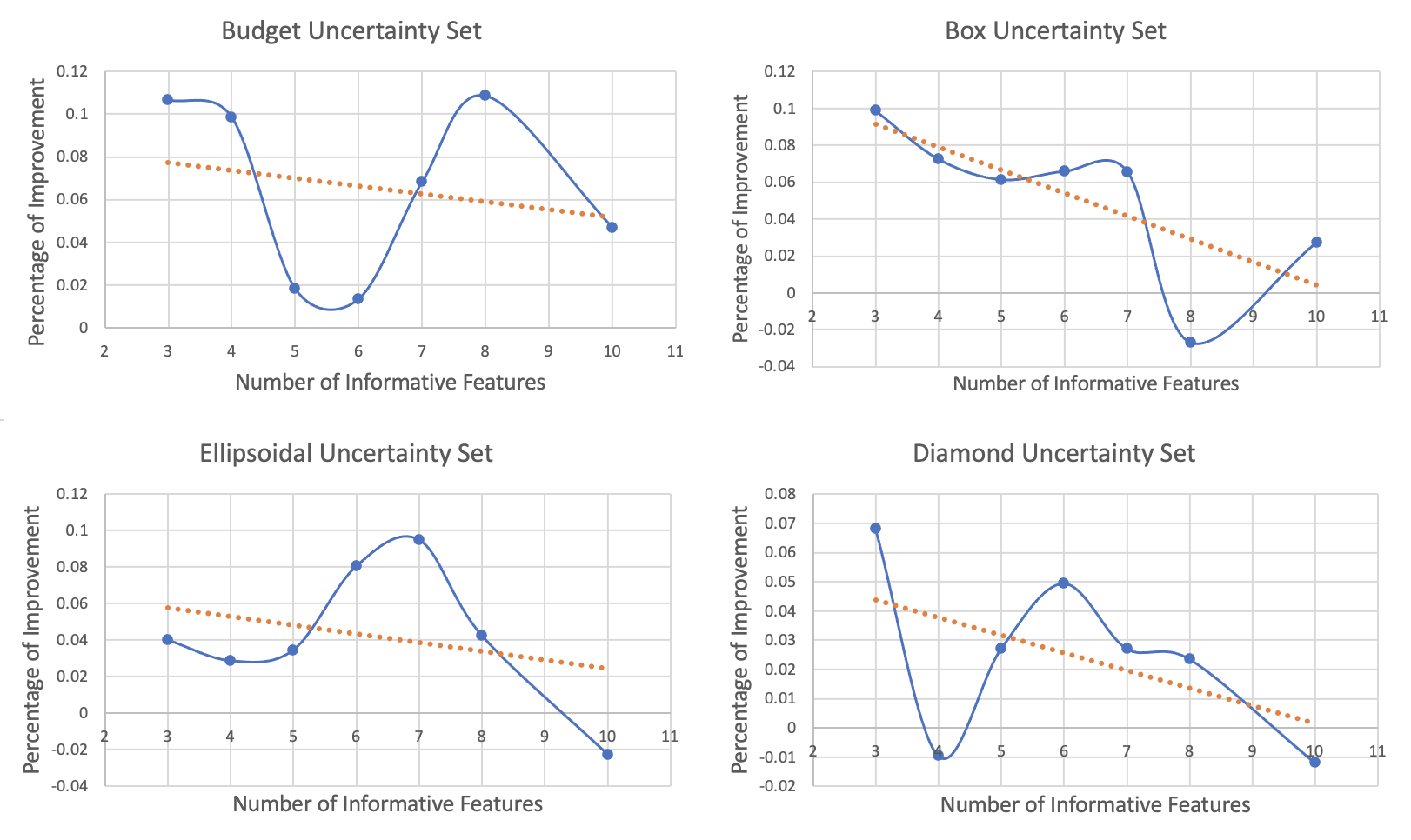}
    \caption{Percentage of improvement of AUR from WUR across different synthetic datasets with different sample sizes and different informative feature sizes. The orange line indicates the trend of improvement, where it monotonically decreases as the sample size increases, and as the number of informative features increases.}
    \label{Fig4}
\end{figure}

\section{Conclusions}
In this work, we have re-considered the nominal robust regression formulation with the worst-case uncertainty set and instead studied the characterizations of the robust regression formulation with averaged uncertainty set. We found that this new formulation establishes the missing connection between the mean squared regression with existing robust regression formulations. More concretely, we found that over all symmetric uncertainty sets we have studied, including the ellipsoidal, box, diamond, budget, and Schatten norm uncertainty sets, the averaged uncertainty formulation is equivalent to the mean squared regression with ridge regularization. We thus established a natural, theoretical connection to the ridge regression under a robust optimization lens. We also show that in the more general, non-symmetric settings of a polytope uncertainty set, this exact equivalence with ridge regression no longer holds.

\hfill 

\noindent
We also justify this formulation as the proper model to solve by evaluating our methodology on both synthetic and real-world datasets and found that empirically, the averaged uncertainty set approach outperforms the worst-case uncertainty case out-of-sample in all experiments. An important observation also lies in the behaviors of the regularization strength selection process, where we observe that the averaged uncertainty approach requires a larger value. However, with adjusted step sizes, the two methods have similar run times in practice.

\hfill 

\noindent
Finally, it should be noted that this new formulation is simple and follows naturally from existing robust optimization formulations and thus can be applied easily to other frameworks. We expect a similar formulation can also be applied to more general settings beyond linear regression, such as matrix regression, robust optimization with solution constraints, as well as discrete robust optimization.

\newpage
\printbibliography

\newpage
\section*{Appendix}
\appendix
\appendix
\section{Proof for Lemma \ref{u1-lemma1}}
\label{u1-proof}
    \begin{align*}
 \int_{\mathcal{U}_1} \bm{\Delta}^T \bm{\Delta} \, \mathrm{d}\mathcal{U}_1
 = \int_{\mathcal{U}_1}
\begin{bmatrix}
\bm{a_1}^T \bm{a_1} & 
\bm{a_1}^T \bm{a_2} & 
\cdots &  
\bm{a_1}^T \bm{a_k} \\
\bm{a_2}^T \bm{a_1} & 
\bm{a_2}^T\bm{a_2} & 
\cdots &  
\bm{a_2}^T \bm{a_k} \\
\cdots & 
\cdots & 
\cdots & 
\cdots \\
\bm{a_k}^T \bm{a_1} & 
\bm{a_k}^T \bm{a_2} & 
\cdots &  
\bm{a_k}^T \bm{a_k}\\
\end{bmatrix} \, \mathrm{d}\mathcal{U}_1.
\end{align*}
\noindent
All entries of this matrix except those on the diagonal are polynomials of elements of $\bm\Delta$ with exponent 1. Thus, using Lemma \ref{ellip-1}, this expression can be simplified to be the following: 
\begin{align*}
 \int_{\mathcal{U}_1} \bm{\Delta}^T \bm{\Delta} \, \mathrm{d}\mathcal{U}_1 
 = \int_{\mathcal{U}_1}
\begin{bmatrix}
\bm{a_1}^T \bm{a_1} & 0 & \cdots & 0 \\
0 & \bm{a_2}^T\bm{a_2} & \cdots &  0 \\
\cdots & \cdots & \cdots & \cdots \\
0 & 0 & \cdots &  \bm{a_k}^T\bm{a_k}\\
\end{bmatrix} \, \mathrm{d}\mathcal{U}_1.
\end{align*}
\noindent
By symmetry, we also have that $\int_{\mathcal{U}_1} \bm{a_1}^T\bm{a_1} \, \mathrm{d}\mathcal{U}_1 = \int_{\mathcal{U}_1} \bm{a_2}^T\bm{a_2} \, \mathrm{d}\mathcal{U}_1 = \cdots = \int_{\mathcal{U}_1} \bm{a_k}^T\bm{a_k} \, \mathrm{d}\mathcal{U}_1 = \frac{V(nk, \rho)}{k}$, and thus
\begin{align*}
\int_{\mathcal{U}_1} \bm{\Delta}^T \bm{\Delta} \, \mathrm{d}\mathcal{U}_1 = 
\begin{bmatrix}
\frac{V(nk, \rho)}{k} & 0 & \cdots & 0 \\
0 & \frac{V(nk, \rho)}{k} & \cdots &  0 \\
\cdots & \cdots & \cdots & \cdots \\
0 & 0 & \cdots & \frac{V(nk, \rho)}{k}
\end{bmatrix}. & \qedhere
\end{align*}

\section{Proof for Lemma \ref{u2-lemma1}}
\label{u2-proof}

\noindent
We will first show that given $\bm{x} \in \mathbb{R}^n$, and $x_i$ being a component of the vector $\bm{x}$, we have the following: 
    \[
    \int_{\mathcal{U}_2} x_i^2 \, \mathrm{d}\mathcal{U}_2= \frac{(2\rho)^n \rho}{3} \]
\noindent
Without loss of generality, we consider $x_i = x_n$. 
    \begin{align*}
    & \int_{\mathcal{U}_2} x_n^2 \, \mathrm{d}\mathcal{U}_2\\
    & \int_{\| \bm{x} \|_{F_{\infty}} \leq \rho} x_n^2 \, \mathrm{d}\mathcal{U}_2 \\
    &= \underbrace{\int_{-\rho}^{\rho} \cdots \int_{-\rho}^{\rho}}_{n-1} \int_{-\rho}^{\rho} x_n^2 \, \mathrm{d}x_n \underbrace{\, \mathrm{d}x_1 \cdots \, \mathrm{d}x_{n-1}}_{n-1}\\
    &= \underbrace{\int_{-\rho}^{\rho} \cdots \int_{-\rho}^{\rho}}_{n-1} \frac{2\rho^3}{3} \underbrace{\, \mathrm{d}x_1 \cdots \, \mathrm{d}x_{n-1}}_{n-1}\\
    &= (2\rho)^{n-1}\frac{2\rho^3}{3}\\
    &= \frac{(2\rho)^n \rho^2}{3}.  \qedhere
\end{align*}

\noindent
Applied in the original lemma setting, we have:
\begin{align*}
 \int_{\mathcal{U}_2} \bm{\Delta}^T \bm{\Delta} \, \mathrm{d}\mathcal{U}_2 
 = \int_{\mathcal{U}_2}
\begin{bmatrix}
\bm{a_1}^T \bm{a_1} & 
\bm{a_1}^T \bm{a_2} & 
\cdots &  
\bm{a_1}^T \bm{a_k} \\
\bm{a_2}^T \bm{a_1} & 
\bm{a_2}^T\bm{a_2} & 
\cdots &  
\bm{a_2}^T \bm{a_k} \\
\cdots & 
\cdots & 
\cdots & 
\cdots \\
\bm{a_k}^T \bm{a_1} & 
\bm{a_k}^T \bm{a_2} & 
\cdots &  
\bm{a_k}^T \bm{a_k}\\
\end{bmatrix} \, \mathrm{d}\mathcal{U}_2
\end{align*}
\noindent
where first for the off-diagnoal entries, 
\begin{align*}
    & \int_{\mathcal{U}_2} \bm{a_i}^T \bm{a_j} \, \mathrm{d}\mathcal{U}_2 \\ 
    &= \int_{\mathcal{U}_2} \sum_{\ell=1}^n a_{i\ell} a_{j\ell} \, \mathrm{d}\mathcal{U}_2\\
    &= \sum_{\ell=1}^n \int_{\mathcal{U}_2}  a_{i\ell} a_{j\ell} \, \mathrm{d}\mathcal{U}_2 = 0
\end{align*}
\noindent
for the diagonal entries, 
\begin{align*}
    & \int_{\mathcal{U}_2} \bm{a_i}^T\bm{a_i} \, \mathrm{d}\mathcal{U}_2 \\ 
    &= \int_{\mathcal{U}_2} \sum_{\ell=1}^n a_{i\ell}^2 \, \mathrm{d}\mathcal{U}_2\\
    &= \sum_{\ell=1}^n \int_{\mathcal{U}_2} a_{i\ell}^2 \, \mathrm{d}\mathcal{U}_2\\
    &= \sum_{\ell=1}^n \frac{(2\rho)^{nk} \rho^2}{3} \\
    &= \frac{(2\rho)^{nk} \rho^2 n}{3}  \qedhere
\end{align*}

\section{Proof for Lemma \ref{u3-lemma1}}
\label{u3-proof}
Let $V_{n-1}$ be the volume of the diamond uncertainty set $\mathcal{U}_3$ in the $(n-1)$-th dimension. Let $y_i = \frac{x_i}{\rho - x_n}$, and $z_i = \rho y_i$, without loss of generality, consider $x_i = x_n$. 
\begin{align*}
    & \int_{\mathcal{U}_3} x_n^2 \, \mathrm{d}\mathcal{U}_3 \\
    & \int_{\| \bm{x} \|_{F_1} \leq \rho} x_n^2 \, \mathrm{d}\mathcal{U}_3 \\
    &= 2^n \int_{x_1 + \cdots + x_{n} \leq \rho, x_i \geq 0 \hspace{1mm} \forall i} x_n^2 \, \mathrm{d}\mathcal{U}_3 \\ 
    &= 2^n \int_0^{\rho} \left(\int_{x_1 + \cdots + x_{n-1} \leq \rho - x_n, x_i \geq 0} 1 \, \mathrm{d}x_1 \cdots \mathrm{d}x_{n-1}\right) x_n^2 \, \mathrm{d}x_n)\\
    &= 2^n \int_0^{\rho} \left(\int_{y_1 + \cdots + y_{n-1} \leq 1, y_i \geq 0} (\rho - x_n)^{n-1} \, \mathrm{d}y_1 \cdots \mathrm{d}y_{n-1} \right) x_n^2 \, \mathrm{d}x_n)\\
    &= 2^n \int_0^{\rho} \left(\int_{z_1 + \cdots + z_{n-1} \leq \rho, z_i \geq 0} 
     \frac{(\rho - x_n)^{n-1}}{\rho^{n-1}} \mathrm{d}z_1 \cdots \mathrm{d}z_{n-1} \right) x_n^2 \, \mathrm{d}x_n)\\
    &= 2^n \int_0^{\rho} \left(\int_{z_1 + \cdots + z_{n-1} \leq \rho, z_i \geq 0} 
     (1-\frac{x_n}{\rho})^{n-1}
    \, \mathrm{d}z_1 \cdots \mathrm{d}z_{n-1}\right) x_n^2 \, \mathrm{d}x_n)\\
    &= 2^n \int_0^{\rho} 
     (1-\frac{x_n}{\rho})^{n-1}
    \frac{V_{n-1}}{2^{n-1}}
     x_n^2 \, \mathrm{d}x_n\\
    &= 2 V_{n-1} \int_0^{\rho} 
     (1-\frac{x_n}{\rho})^{n-1}
     x_n^2 \, \mathrm{d}x_n\\
    &= 2 \frac{(2\rho)^{n-1}}{(n-1)!} \frac{2\rho^3}{n(n+1)(n+2)}\\
    &= \frac{(2\rho)^{n+1}\rho}{(n+2)!}.  \qedhere
\end{align*}

\hfill

\noindent
Using the conclusion above, we have the following: 
\begin{align*}
 \int_{\mathcal{U}_3} \bm{\Delta}^T \bm{\Delta} \, \mathrm{d}\mathcal{U}_3 
 = \int_{\mathcal{U}_3}
\begin{bmatrix}
\bm{a_1}^T \bm{a_1} & 
\bm{a_1}^T \bm{a_2} & 
\cdots &  
\bm{a_1}^T \bm{a_k} \\
\bm{a_2}^T \bm{a_1} & 
\bm{a_2}^T \bm{a_2} & 
\cdots &  
\bm{a_2}^T \bm{a_k} \\
\cdots & 
\cdots & 
\cdots & 
\cdots \\
\bm{a_k}^T \bm{a_1} & 
\bm{a_k}^T\bm{a_2} & 
\cdots &  
\bm{a_k}^T\bm{a_k}\\
\end{bmatrix} \, \mathrm{d}\mathcal{U}_3
\end{align*}
We observe that the elements off-diagonal can all be expressed as 
\begin{align*}
    & \int_{\mathcal{U}_3} \bm{a_i}^T \bm{a_j}  \, \mathrm{d}\mathcal{U}_3 \\ 
    &= \int_{\mathcal{U}_3} \sum_{\ell=1}^n a_{i\ell} a_{j\ell}  \, \mathrm{d}\mathcal{U}_3 \\
    &= \sum_{\ell=1}^n \int_{\mathcal{U}_3} a_{i\ell} a_{j
    \ell} \, \mathrm{d}\mathcal{U}_3 \\
    &= 0
\end{align*}
for the terms in the diagonal, we have: 
\begin{align*}
    & \int_{\mathcal{U}_3} \bm{a_i}^T \bm{a_i} \, \mathrm{d}\mathcal{U}_3 \\ 
    &= \int_{\mathcal{U}_3} \sum_{\ell=1}^n a_{i\ell}^2 \, \mathrm{d}\mathcal{U}_3 \\
    &= \sum_{\ell=1}^n \int_{\mathcal{U}_3} a_{i\ell}^2 \, \mathrm{d}\mathcal{U}_3 \\
    &= \frac{(2\rho)^{nk+1}\rho n}{(nk+2)!}  \qedhere
\end{align*} 

\section{Proof for Lemma \ref{u4-lemma1}}
\label{u4-proof}
\noindent
Without loss of generality, we compute the case of $x_i = x_n$. If we assume that all $x_i \geq 0$, then depending on the value of $x_n$, there can be two cases where
$$
x_1 + \cdots + x_n \leq \rho, x_i \leq \Gamma 
\quad \forall i \in [1:n] 
\begin{cases}
0 \leq x_n \leq \rho - \Gamma, & \text{Case 1, denote as region $A_n$}\\
\rho - \Gamma \leq x_n \leq \Gamma, & \text{Case 2, denote as region $B_n$}
\end{cases}
$$

\hfill 

\noindent
In Case 1, where $y_i = \frac{x_i}{\rho - x_n}$, $z_i = \rho y_i$, and $V_{n-1}$ is the volume defined by Lemma \ref{u4-lem2}, 
\begin{align*}
& \int_{\mathcal{U}_4} x_n^2 \, \mathrm{d} A_n \\
& \int_{\| \bm{x} \|_1 \leq \rho, \| \bm{x} \|_{\infty} \leq \Gamma, 0 \leq x_n \leq \rho - \Gamma} x_n^2 \, \mathrm{d} A_n \\
&= 2^n\int_{0}^{\rho - \Gamma} \int_{x_1 + \cdots + x_n \leq \rho, 0 \leq x_i \leq \Gamma \hspace{1mm} \forall i \in [1:n-1]} x_n^2 \, \mathrm{d}A_{n-1} \\
&= 2^n\int_{0}^{\rho - \Gamma} \int_{x_1 + \cdots + x_{n-1} \leq \rho - x_n, 0 \leq x_i \leq \Gamma \hspace{1mm} \forall i \in [1:n-1]} x_n^2 \, \mathrm{d}x_1 \cdots \mathrm{d}x_{n-1}  \\
&= 2^n\int_{0}^{\rho - \Gamma} \int_{y_1 + \cdots + y_{n-1} \leq 1, 0 \leq y_i \leq \frac{\Gamma}{\rho - x_n} \hspace{1mm} \forall i \in [1:n-1]} (\rho - x_n)^{n-1}x_n^2 \mathrm{d}y_1 \cdots \, \mathrm{d}y_{n-1} \\
&= 2^n\int_{0}^{\rho - \Gamma} \int_{z_1 + \cdots + z_{n-1} \leq 1, 0 \leq z_i \leq \frac{\rho \Gamma}{\rho - x_n} \hspace{1mm} \forall i \in [1:n-1]} \frac{(\rho - x_n)^{n-1}x_n^2}{\rho^{n-1}} \mathrm{d}z_1 \cdots \, \mathrm{d}z_{n-1} \\
&= 2^n\int_{0}^{\rho - \Gamma} \frac{(\rho - x_n)^{n-1}x_n^2}{\rho^{n-1}} \frac{V_{n-1}}{2^{n-1}} \, \mathrm{d}x_n \\
&= 2\int_{0}^{\rho - \Gamma} \frac{(\rho - x_n)^{n-1}x_n^2}{\rho^{n-1}} \frac{(2\rho)^n - (n-1)(2(\rho - \frac{\rho \Gamma}{\rho - x_n}))^{n-1}}{(n-1)!} \, \mathrm{d}x_n \\
&= \frac{2^n}{(n-1)!} \int_{0}^{\rho - \Gamma} \frac{(\rho - x_n)^{n-1}x_n^2}{\rho^{n-1}} \rho^{n-1} \left( 1 - (n-1)(\frac{\rho - \Gamma - x_n}{\rho - x_n})^{n-1} \right) \, \mathrm{d}x_n\\
&= \frac{2^n}{(n-1)!} \int_{0}^{\rho - \Gamma}(\rho - x_n)^{n-1}x_n^2 \frac{(\rho - x_n)^{n-1} - (n-1)(\rho - \Gamma - x_n)^{n-1}}{(\rho - x_n)^{n-1}} \, \mathrm{d}x_n\\
&= \frac{2^n}{(n-1)!} \int_{0}^{\rho - \Gamma}(\rho - x_n)^{n-1}x_n^2 - (n-1)(\rho - \Gamma - x_n)^{n-1}x_n^2 \, \mathrm{d}x_n \\
&= \frac{2^n}{(n+2)!} (\Gamma^n((4n^2+2n)\rho \Gamma - (n^2+n)\Gamma^2 - (n^2 + 3n + 2)\rho^2) \\ & + 2\rho^{n+2} + (\rho - \Gamma)^n ((2-2n)\Gamma^2 + (4n-4)\rho\Gamma + (2-2n)\rho^2))\\
\end{align*}

\noindent
In Case 2, since we have $\rho - \Gamma \leq x_n \leq \Gamma$, we can rewrite $x_1 + \cdots + x_{n-1} \leq \rho - x_n$ as $x_1 + \cdots + x_{n-1} \leq \Gamma$. This implies that in case 2, $x_i \leq \Gamma \hspace{2mm} \forall i \in [1:n-1]$ will be automatically satisfied. We thus instead are dealing with the problem of $x_1 + \cdots + x_{n-1} \leq \rho - x_n, x_i \geq 0$. We recognize that this is exactly the diamond uncertainty set case. Thus we have the following: 
\begin{align*}
& \int_{\mathcal{U}_4} x_n^2 \, \mathrm{d}B_n  \\
& \int_{\| \bm{x} \|_1 \leq \rho, \| \bm{x} \|_{\infty} \leq \Gamma, \rho - \Gamma \leq x_n \leq \Gamma} x_n^2 \, \mathrm{d}B_n  \\
&= 2^n\int_{\rho - \Gamma}^{\Gamma} \int_{x_1 + \cdots + x_n \leq \rho, x_i \geq 0, \hspace{1mm} \forall i \in [1:n-1]} x_n^2 \, \mathrm{d}B_{n-1}  \\
&= 2^n \int_{\rho - \Gamma}^{\Gamma} \left(1-\frac{x_n}{\rho}\right)^{n-1} \frac{V_{n-1}}{2^{n-1}} x_n^2 \, \mathrm{d}x_n \\
&= 2 \int_{\rho - \Gamma}^{\Gamma} \left(1-\frac{x_n}{\rho}\right)^{n-1}\frac{(2\rho)^{n-1}}{(n-1)!} x_n^2 \, \mathrm{d}x_n \\
&= \frac{2^n}{(n-1)!}\int_{\rho - \Gamma}^{\Gamma}(\rho-x_n)^{n-1} x_n^2 \, \mathrm{d}x_n \\
&= \frac{2^n}{(n+2)!} (\Gamma^n((-2n^2 - 4n)\rho\Gamma + (n^2 + n)\Gamma^2 + (n^2 + 3n + 2)\rho^2) + \\ & (\rho - \Gamma)^n ((-n^2 - n)\Gamma^2 - 2n\rho\Gamma - 2\rho^2)). \qedhere
\end{align*}

\hfill 

\noindent
Putting everything together, we have that, 
\begin{align*}
& \int_{\| \bm{x} \|_1 \leq \rho, \| \bm{x} \|_{\infty} \leq \Gamma} x_n^2 \, \mathrm{d}\mathcal{U}_4 \\
&= \int_{\| \bm{x} \|_1 \leq \rho, \| \bm{x} \|_{\infty} \leq \Gamma, 0 \leq x_n \leq \rho - \Gamma} x_n^2 \, \mathrm{d}A_n + \int_{\| \bm{x} \|_1 \leq \rho, \| \bm{x} \|_{\infty} \leq \Gamma, \rho - \Gamma \leq x_n \leq \Gamma} x_n^2 \, \mathrm{d}B_n\\
&= \frac{2^n}{(n+2)!} (2\rho^{n+2} - (\rho - \Gamma)^n ((n^2 + 3n - 2)\Gamma^2 + (4-2n)\rho \Gamma + 2n\rho^2) \\
&= \frac{2\rho^2}{(n+1)(n+2)}\frac{(2\rho)^n - n(2(\rho - \Gamma))^n}{n!} - \frac{(2(\rho - \Gamma))^n}{n!}\frac{(n^2+3n-2)\Gamma^2 + (4-2n)\rho \Gamma}{(n+1)(n+2)}. \qedhere
\end{align*}

\end{document}